\def\paperID{10133}
\def\confName{CVPR}
\def\confYear{2024}

\def\authorBlock{
    Zihan Gao \qquad
    Licheng Jiao \thanks{Corresponding author. This work was supported in part by
the Key Scientific Technological Innovation Research Project by Ministry of Education,
the Joint Funds of the National Natural Science Foundation of China (U22B2054),
the National Natural Science Foundation of China (62076192, 61902298, 61573267, 61906150, and 62276199),
the 111 Project,
the Program for Cheung Kong Scholars and Innovative Research Team in University (IRT 15R53),
the ST Innovation Project from the Chinese Ministry of Education,
the Key Research and Development Program in Shaanxi Province of China(2019ZDLGY03-06),
the China Postdoctoral fund(2022T150506).} \qquad
    Lingling Li \qquad
    Xu Liu \qquad
    Fang Liu \\
    Puhua Chen \qquad 
    Yuwei Guo \\
    School of Artificial Intelligence, Xidian University \\
}

\newif\ifreview 

\newif\ifarxiv 
\newcommand{\arxiv}{\arxivtrue}
\newif\ifcamera 

\newif\ifrebuttal 

\arxiv
\pdfoutput=1
\documentclass[10pt,twocolumn,letterpaper]{article}
\ifreview \usepackage[review]{cvpr} \fi
\ifarxiv \usepackage[pagenumbers]{cvpr} \fi
\ifrebuttal \usepackage[rebuttal]{cvpr} \fi
\ifcamera \usepackage{cvpr} \fi

\usepackage{graphicx}	
\usepackage{amsmath}	
\usepackage{amssymb}	
\usepackage{booktabs}
\usepackage{times}
\usepackage{microtype}
\usepackage{epsfig}
\usepackage[table,xcdraw,dvipsnames]{xcolor}
\usepackage{caption}
\usepackage{float}
\usepackage{placeins}
\usepackage{color, colortbl}
\usepackage{stfloats}
\usepackage{enumitem}
\usepackage{tabularx}
\usepackage{xstring}
\usepackage{multirow}
\usepackage{xspace}
\usepackage{url}
\usepackage{subcaption}
\usepackage{xcolor}
\usepackage{adjustbox}
\usepackage[hang,flushmargin]{footmisc}
\usepackage{tikz}
\ifcamera \usepackage[accsupp]{axessibility} \fi

\definecolor{bestcolor}{rgb}{ .98,  .706,  .694}
\definecolor{secondcolor}{rgb}{1, .863, .769}
\definecolor{thirdcolor}{rgb}{ 1,  .976,  .89}

% ...

\ifarxiv  \fi

\newcommand{\R}[1]{{%
    \textbf{%
        \ifstrequal{#1}{1}{\textcolor{red}{R#1}}{%
        \ifstrequal{#1}{2}{\textcolor{blue}{R#1}}{%
        \ifstrequal{#1}{3}{\textcolor{magenta}{R#1}}{%
        \ifstrequal{#1}{4}{\textcolor{teal}{R#1}}{%
                           \textcolor{cyan}{R#1}%
        }}}}%
    }%
}}

\usepackage{xr-hyper}

\makeatletter
\newcommand*{\addFileDependency}[1]{
  \typeout{(#1)}
  \@addtofilelist{#1}
  \IfFileExists{#1}{}{\typeout{No file #1.}}
}

\makeatother

\definecolor{cvprblue}{rgb}{0.21,0.49,0.74}
\usepackage[pagebackref,breaklinks,colorlinks,citecolor=cvprblue]{hyperref}
\usepackage[capitalize]{cleveref}
\crefname{section}{Sec.}{Secs.}
\crefname{table}{Table}{Tables}
\crefname{figure}{Fig.}{Figs.}

\begin{document}
%% TITLE
\title{Multiplane Prior Guided Few-Shot Aerial Scene Rendering}
\author{\authorBlock}

\maketitle
\begin{abstract}

Neural Radiance Fields (NeRF) have been successfully applied in various aerial scenes, yet they face challenges with sparse views due to limited supervision. The acquisition of dense aerial views is often prohibitive, as unmanned aerial vehicles (UAVs) may encounter constraints in perspective range and energy constraints. In this work, we introduce Multiplane Prior guided NeRF (MPNeRF), a novel approach tailored for few-shot aerial scene rendering—marking a pioneering effort in this domain. Our key insight is that the intrinsic geometric regularities specific to aerial imagery could be leveraged to enhance NeRF in sparse aerial scenes. By investigating NeRF's and Multiplane Image (MPI)'s behavior, we propose to guide the training process of NeRF with a Multiplane Prior. The proposed Multiplane Prior draws upon MPI's benefits and incorporates advanced image comprehension through a SwinV2 Transformer, pre-trained via SimMIM. Our extensive experiments demonstrate that MPNeRF outperforms existing state-of-the-art methods applied in non-aerial contexts, by tripling the performance in SSIM and LPIPS even with three views available. We hope our work offers insights into the development of NeRF-based applications in aerial scenes with limited data.

\end{abstract}
\section{Introduction}
\label{sec:intro}

Neural Radiance Field (NeRF) \cite{mildenhall2020nerf} has succeeded in rendering high-fidelity novel views and many 3D applications by modeling 3D scenes as a continuous implicit function. In contrast to indoor or synthetic scenes capturing simple objects using cell phones, aerial images provide a unique bird's-eye view and overview of a landscape.
Based on NeRF, many applications in aerial scenes have been developed, such as navigation, urban planning, data augmentation, autonomous vehicles, and environmental mapping \cite{de2022next,alvey2021simulated,maxey2023uav,sucar2021imap,zhu2022nice,kwon2023renderable,maggio2023loc}.

In many real-world scenarios, unmanned aerial vehicles (UAVs) encounter constraints such as limited perspectives, energy limitations, or adverse weather conditions, which restrict their ability to acquire dense observational data. While NeRF forms a foundational technology for many aerial applications, it is prone to overfitting on sparse training views \cite{niemeyer2022regnerf, kim2022infonerf,jain2021putting}. This limitation of NeRF becomes particularly salient in the context of aerial imagery. Alleviating this problem could save resources and may benefit numerous applications.

As the first to explore few-shot NeRF for aerial imagery, we stand at the forefront of this emerging field. 
The landscape of few-shot neural rendering to date has been predominantly shaped by its application to indoor and synthetic scenes.
Transfer learning based methods \cite{yu2021pixelnerf, chen2021mvsnerf, wang2021ibrnet, wang2022attention} aim to pre-train the model on a large number of scenes. 
Yet, these approaches necessitate extensive datasets for pre-training. This is not only resource-heavy but also impractical to fulfill for varied aerial scenarios.
Another line of works \cite{kim2022infonerf, niemeyer2022regnerf, ehret2022regularization, jain2021putting, xu2022sinnerf, yang2023freenerf, Kanaoka_2023_BMVC, seo2023flipnerf} seeks to impose regularization on NeRF by exploring the universal attributes of 3D scenes like local continuity and semantic consistency. Yet, in situations where available data is significantly sparse relative to the complexity of the scene, these methods might struggle to maintain their effectiveness.
And the last is depth-prior-based methods \cite{deng2022depth, wang2023sparsenerf} gain additional supervision from the scene's depth. These methods can be problematic in aerial images due to the frequent occurrence of ambiguous depth cues and the high cost of obtaining accurate depth maps.
Despite their efficacy in controlled environments, these methods fall short of addressing the unique complexities of aerial scenes, leaving a gap that our work aims to fill. We therefore ask: \textbf{Can we harness the intrinsic geometric regularities specific to aerial imagery to broaden the capabilities of NeRF under sparse data conditions, thereby easing the data collection constraints? }

To answer this question, we turn to earlier works on 2.5D representations such as Multiplane Image (MPI) \cite{zhou2018stereo, tucker2020single, li2021mine, wu2022remote}. MPI typically operates by extracting multiple RGB and density planes from a single image input by an encoder-decoder style MPI generator to represent scene geometry within the camera's frustum. Although NeRF provides a continuous representation of a scene, MPI offers discrete, frustum-confined layers that can be particularly advantageous in the context of aerial imagery. This is due to UAVs frequently capturing images from overhead perspectives that align well with MPI's planar representation. Additionally, the encoder-decoder architecture of the MPI generator can exploit the inductive biases inherent in advanced convolutional and self-attention-based image processing compared to the simple multi-layer perceptron (MLP) of NeRF, thus enhancing the rendering of local and global scene details. However, while MPIs present certain benefits in terms of their adaptability to aerial perspectives, their partial scene recovery and limitation to individual frustums pose challenges in creating a comprehensive 3D understanding. 

In this work, we present Multiplane Prior guided NeRF (MPNeRF), a novel method for enhancing NeRF models in few-shot aerial scene rendering. We guide NeRF's learning process by using a multiplane prior—a concept drawn from the strengths of MPI and refined with cutting-edge image understanding from a Swin TransformerV2 pre-trained with SimMIM \cite{liu2022swin}. This approach unites the capabilities of NeRF with the perspective-friendly nature of MPI, tailored for the unique vantage points of aerial scenes. Concretely, our approach updates the NeRF branch using pseudo labels generated from the MPI branch. As training proceeds, NeRF can effectively pick up finer details from the MPI branch and the advantage of the MPI branch is implicitly distilled into NeRF. This strategy implicitly folds a multiplane prior to NeRF, boosting its performance in handling sparse aerial imagery data.
Our contributions can be summarized as follows:
\begin{enumerate}
\item We introduce Miltiplane Prior guided NeRF (MPNeRF), a novel framework that synergistically combines NeRF and MPIs for enhanced few-shot neural rendering in aerial scenes. To the best of our knowledge, this is the first method specially designed for this task.
\item Through an investigation, we pinpoint and analyze the typical failure modes of NeRF and MPI in aerial scenes. We devise a simple yet effective learning strategy that guides the training process of NeRF by learning a multiplane prior, effectively circumventing NeRF's typical pitfalls in sparse aerial scenes.
\item We compare MPNeRF against a suite of state-of-the-art non-aerial scene methods, rigorously testing its adaptability and performance in aerial scenarios. Our experiments demonstrate MPNeRF's superior performance, showcasing its significant leap over methods previously confined to non-aerial contexts.
\end{enumerate}

\section{Related Work}
\label{sec:related}

\subsection{Scene Representations for View Synthesis.}
Earlier works on light fields \cite{Levoy1996LightFR, 10.1145/3596711.3596760, davis2012unstructured} achieve view synthesis by interpolating nearby views given a dense set of input images. Later works utilize explicit mesh \cite{waechter2014let, li2018differentiable, liu2019soft, chen2019learning}, or volumetric \cite{kutulakos2000theory, seitz1999photorealistic, kar2017learning, henzler2020learning, sitzmann2019deepvoxels, wiles2020synsin} representation to represent the scene.
More recently, layered representations have gained attention due to their efficiency in modeling occluded content. 
One such layered representation is the MPI \cite{zhou2018stereo, tucker2020single, li2021mine, wu2022remote}. An MPI consists of multiple planes of RGB and $\alpha$ values at fixed depths. Given an input image, an encoder-decoder network typically generates the MPI within the camera frustum. This MPI is then homography warped to the target camera position and integrated over the planes to produce novel views. It's important to note that the generated MPI only models the geometry within each camera frustum at given depths, and the complete 3D scene is not fully recovered.

Recently, NeRF \cite{mildenhall2020nerf} has shown significant potential in novel view synthesis. NeRF works by modeling the scene with a continuous function of 3D coordinates and viewing directions to output the corresponding RGB and volume density values. Following NeRF, many methods have been proposed. mip-NeRF \cite{barron2021mip} introduces a more robust representation that uses a cone tracing technique and samples the cone with multivariate Gaussian.
NeRF-W \cite{RicardoMartinBrualla2020NeRFIT} and Ha-NeRF\cite{chen2022hallucinated} have extended the applicability of NeRF to in-the-wild photo collections through object decomposition and hallucination techniques. For large-scale scenes, BungeeNeRF \cite{YuanboXiangli2021singleNeRFPN} proposes a multiscale progressive learning method to reconstruct cities from satellite imagery, while Block-NeRF \cite{tancik2022block} leverages individual NeRFs for each component of the scene to achieve large-scale scene rendering. ShadowNeRF \cite{DawaDerksen2021ShadowNR} and Sat-NeRF \cite{RogerMari2023SatNeRFLM} address the issue of strong noncorrelation between satellite images taken at different times by modeling solar light and transient objects. However, these NeRF-based methods are still limited by the need for densely sampled views of the scene. 

\subsection{NeRF with Sparse Input}

Many approaches have been developed to train a NeRF from sparse input in different directions. One straightforward direction \cite{yu2021pixelnerf, chen2021mvsnerf, wang2021ibrnet, trevithick2020grf, liu2022neural, wang2022attention} is to learn the general appearance of a scene or object from a large number of data. PixelNeRF \cite{yu2021pixelnerf} adopts a CNN feature extractor to condition each input coordinate with image features. MVSNeRF \cite{chen2021mvsnerf} uses 3D CNN to process cost volume acquired by image warping. These methods often require a large number of multi-view images to be pre-trained on, which is sometimes hard to acquire in aerial imagery. Some other techniques \cite{kim2022infonerf, niemeyer2022regnerf, ehret2022regularization, kwak2023geconerf, seo2023flipnerf} find it is more data-efficient to regularize NeRF with common properties of the 3D geometry. InfoNeRF \cite{kim2022infonerf} regularizes NeRF by putting a sparsity constraint on the density of each ray. RegNeRF \cite{niemeyer2022regnerf} regularizes NeRF by local smoothness.  Other works \cite{deng2022depth, jain2021putting, wei2021nerfingmvs, wang2023sparsenerf} aim to take advantage of supervision from other sources, such as depth or appearance. DS-NeRF \cite{deng2022depth} supervises the geometry with sparse point cloud generated with structure from motion. DietNeRF \cite{jain2021putting} regularize NeRF by ensuring perceptual consistency within different views. ManifoldNeRF \cite{Kanaoka_2023_BMVC} builds upon DietNeRF and takes into account viewpoint-dependent perceptual consistency to refine supervision in unknown viewpoints.
However, we noticed none of these methods is designed for aerial scenes and thus left a gap our work aims to fill.

\section{Method}
\label{sec:method}
\begin{figure*}
    \centering
    \includegraphics[width=1\linewidth]{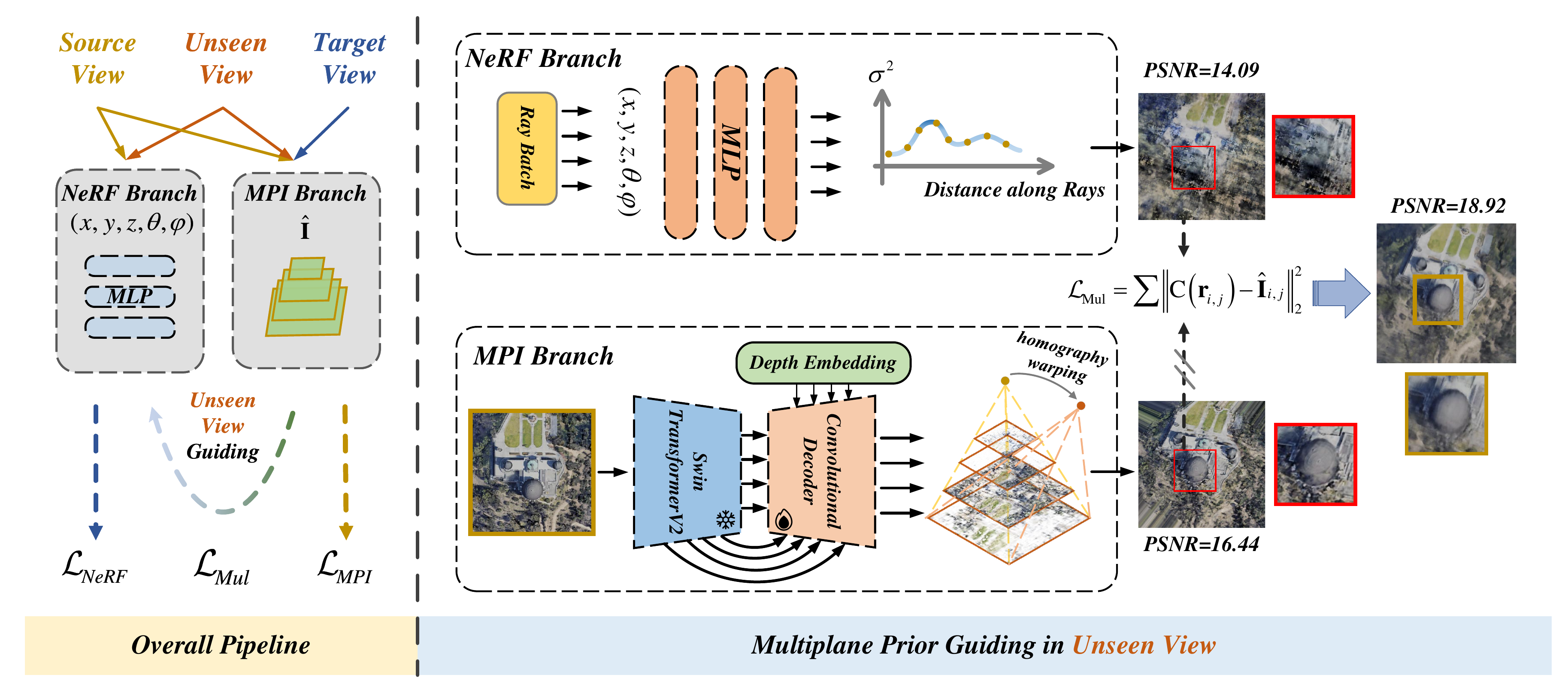}
    \caption{\textbf{Overall pipeline for training Multiplane Prior guided NeRF (MPNeRF).} Our novel MPNeRF architecture integrates a standard NeRF branch with an MPI branch, informed by a pre-trained SwinV2 Transformer. This design introduces a multiplane prior to guide the NeRF training, addressing the common challenges of rendering with sparse aerial data. The process begins by sampling three distinct views: a source and target view for training with known ground truth, and an unseen view from a novel viewpoint. The NeRF model is then refined using pseudo labels produced by the MPI branch, which are especially crucial for synthesizing views from previously unseen angles, as shown in the pipeline. }
    \label{fig:overall}
\end{figure*}

Our objective is to train a standard NeRF model to create highly realistic novel views of an aerial scene from a limited number of captured perspectives. To address the challenges of training NeRF with sparse aerial views, we introduce a novel training approach that leverages a Multiplane Prior. The proposed Multiplane Prior harnesses the strengths of MPI and is enriched by advanced image understanding capabilities derived from a SwinV2 Transformer pre-trained using SimMIM \cite{xie2022simmim}. An overview of our approach is presented in Figure \ref{fig:overall}.

In Sec. \ref{sec:preliminaries}, we briefly review the background related to our method. Sec. \ref{sec:analysis} provides an investigation into NeRF and MPI's behavior in sparse aerial scenes. Sec. \ref{sec:framework} describe our overall framework.

\subsection{Preliminaries}
\label{sec:preliminaries}
\noindent \textbf{Neural Radiance Field.}
Given a 3D coordinate $\mathbf{x}=\left(x,y,z\right)$ and a 2D viewing direction $\mathbf{d}=\left(\theta,\varphi \right)$, NeRF aims to model the scene by solving a continuous function $f\left(\mathbf{x}, \mathbf{d}\right)=\left(\mathbf{c},\sigma\right)$ using multi-layer perceptron (MLP) network, where $\mathbf{c}$ and $\sigma$ represent the emissive color and volume density at the given coordinate. NeRF cast rays $\mathbf{r}\left(t\right)=\mathbf{o}+t\mathbf{d}$ from the camera origin $\mathbf{o}$ along the direction $\mathbf{d}$ to pass through a pixel. NeRF then samples $M$ points along this ray and computes its color by volume rendering:
\begin{equation}
\begin{aligned}
 &\hat{\operatorname{C}}\left(\mathbf{r}\right)={\sum_{i=1}^{M}} T_{i}\left(1-\exp \left(-\sigma_{i} \delta_{i}\right)\right) \mathbf{c}_{i}, \\ 
 &T_{i}=\exp \left(-\sum_{j=1}^{i-1} \sigma_{j} \delta_{j}\right),
\end{aligned}
\end{equation}
where $\mathbf{c}_{i}$ and $\sigma_{i}$ are the color and volume density of i-th sample along the ray and $\delta_{i}$ is the distance between adjacent samples. $\operatorname{\hat{C}}\left(\mathbf{r}\right)$ denotes the final color of that pixel rendered by NeRF. 
% Subsequently, the scene depth for a given pixel can be obtained by:
% \begin{equation}
%     \hat{\operatorname{d}}(\mathbf{r})={\sum_{i=1}^{M}} T_{i}\left(1-\exp \left(-\sigma_{i} \delta_{i}\right)\right) t,
% \end{equation}
In NeRF, a dense 3D scene is recovered implicitly in the form of neural network weights.

\noindent \textbf{Multiplane Image.}
Multi-plane Image (MPI) represents the scene by dividing the 3D space into a collection of planes with RGB and density values in one camera frustum. 
In training, each batch consists of a pair of images $\mathbf{I_{s}},\mathbf{I_{t}}\in\mathbb{R}^{H\times W \times 3}$ with corresponding camera intrinsic $\mathbf{K_{s}},\mathbf{K_{t}}\in\mathbb{R}^{3 \times 3}$ and relative pose $\mathbf{P}_{s2t}=\left[ \mathbf{R}_{s2t} \in \mathbb{R}^{3 \times 3}, \mathbf{t}_{s2t} \in \mathbb{R}^{3}  \right]$ denoted as $\left\{\left(\mathbf{I}_{s},\mathbf{K}_{s}\right),\left(\mathbf{I}_{t},\mathbf{K}_{t},\right), \mathbf{P}\right\}$, subscripts $s$ and $t$ represent source and target viewpoint respectively. Depth for each plane is sampled $\{z=z_{k} | k=1,2,3,\cdots ,N \}$ uniformly according to the scene bounds. An encoder-decoder based MPI-generator denoted as $\operatorname{G}_{\text{MPI}}$ is adopted to generate multiple planes of RGB and density at discrete depth as:
\begin{equation}
\begin{aligned}
\left\{\left(\mathbf{c}_{{k}},\sigma_{{k}}\right)  | k=1,2,3,\cdots ,N\right\} = \operatorname{G}_{\text{MPI}}\left(\mathbf{I}_{s}\right).
\end{aligned}
\end{equation}
Here the subscript ${k}$ denotes the k-th plane. The rendering of MPI is performed in two steps: First, establish the correspondence between the pixel coordinates in the source and target plane through homography warping as:
\begin{equation}
\begin{aligned}
\left[ u_{t},v_{t},1 \right ]^{T} = \mathbf{K}_{t}\left(\mathbf{R}_{s2t}-\frac{\mathbf{t}_{s2t} \mathbf{n}^{T}}{z_{k}}\right) \mathbf{K}_{s}^{-1}\left [ u_{s},v_{s},1 \right ]^{T}.
\end{aligned}
\end{equation}
Second, similar to NeRF, apply differentiable rendering to get target view 2D images $\hat{\mathbf{I}}_{t}$ :
\begin{equation}
\begin{aligned}
 &\hat{\mathbf{I}}_{t}=\sum_{k=1}^{N} T_{k}\left(1-\exp \left(-\sigma_{k} \delta_{k}\right)\right) \mathbf{c}_{k}, \\
 % &\hat{\mathbf{D}}_{t}=\sum_{i=1}^{N} T_{i}\left(1-\exp \left(-\sigma_{z_{i}} \delta_{z_{i}}\right)\right) {z_{i}}, \\
 &T_{i}=\exp \left(-\sum_{l=1}^{k-1} \sigma_{l} \delta_{l}\right).
\end{aligned}
\end{equation}

\subsection{A Closer Look at The Behavior of NeRF \& MPI}
\label{sec:analysis}

To better understand the behavior of NeRF and MPI, we conducted an investigation into their failure modes. In Figure \ref{fig:ana}, we visualize the rendering results of NeRF and MPI when encountering large camera movements.  
Our findings reveal that NeRF often produces blurry renderings, while MPI tends to exhibit overlapping ghosting effects and cropped corners.

Recall that NeRF represents the whole 3D scene continuously by encoding the volume density and color into an 8-layer MLP's weights. In other words, NeRF utilizes a learning-based approach by forcing correct rendering from every angle of the scene with multi-view consistency. 
Such a model is highly compact when supervised with sufficient training views. When the supervised angle is limited, areas covered less (as in the non-overlapped camera frustum in Figure \ref{fig:ana} (b)) are uncontrolled and may exhibit high-density values \cite{niemeyer2022regnerf, yang2023freenerf, li2023regularize}, leading to blurry or even collapsed results. 
Considering the structured nature of 3D aerial scenes, we recognize two key factors of aerial scenes: \textit{aligned perspectives with predominant planarity}, and \textit{consistent geometric appearance}. First, the typical flight paths of UAVs over these scenes predominantly capture landscapes aligned with the XY planes, providing a unique geometrical consistency. Second, objects in aerial scenes contain common visual characteristics, offering additional cues for scene interpretation and analysis.

In contrast to NeRF, MPI models the scene within each camera frustum and decomposes it into an explicit set of discrete 2D planes at fixed depths. 
This mirrors the overhead views and planar surfaces commonly found in aerial scenes. Also, the convolution-based or self-attention-based MPI-generator is inherently suited to carry prior knowledge of the scene. However, with insufficient supervision provided, the MPI for different camera frustums may not be properly calibrated. As a consequence, we observe the occurrence of overlapping ghosting effects in rendered unseen views. Additionally, when there is substantial camera movement, the corners of the target views may be excluded from the source views, resulting in invalid renderings. However, MPI is successful in preserving high-frequency details in the rendered image. We attribute this capability to the power of CNNs and the implicit encoding of prior knowledge in the MPI generator.

Due to the distinct failure mode of MPI and its favorable properties, we explore a strategy to enhance NeRF in a sparse aerial context.

\subsection{Guiding NeRF with a Multiplane Prior}
\label{sec:framework}
Based on the investigation of the different properties shown in NeRF and MPI when encountering sparse input. We turn to the task of few-shot aerial scene rendering and propose a simple yet effective strategy that treats the MPI as a bridge to convey information that is hard to learn by the traditional NeRF pipeline. 

We formulate the proposed MPNeRF with a NeRF branch, and an MPI branch denoted as $\mathbf{G}_{\theta_{1}}$ and $\mathbf{G}_{\theta_{2}}$. Given a batch contains images from source and target viewpoints alongside the corresponding camera parameters. To train the NeRF branch, we cast rays for the source viewpoints pixels using the camera parameters following \cite{mildenhall2020nerf}. The MSE loss is adopted to supervise the NeRF branch with the ground truth color $\operatorname{C}\left(\mathbf{r}\right)$:
\begin{equation}
\begin{aligned}
\mathcal{L}_{\text{NeRF}}=\sum_{\mathbf{r}\in \mathbf{B}}  {\left \| \hat{\operatorname{C}}\left(\mathbf{r}\right) - \operatorname{C}\left(\mathbf{r}\right) \right \| _{2}^{2}},
\label{eq:4}
\end{aligned}
\end{equation}
where, $B$ is the set of input rays during training.
For the MPI branch, the encoder-decoder style MPI generator takes in images from the source view and outputs the corresponding MPI representation. In order to incorporate prior knowledge, we adopt a frozen Swin Transformer V2 model pre-trained with SimMIM \cite{liu2022swin} as a feature extractor to extract multi-scale features from aerial images. These features are fused to generate the final MPI representation.
The loss function to optimize the MPI branch contains three components: L1 loss to match the synthesized target image $\hat{\mathbf{I}}_{t}$ to ground truth $\mathbf{I}_{t}$ at a pixel level, SSIM loss to encourage structure consistency, and LPIPS loss \cite{zhang2018unreasonable} for perceptual consistency.
\begin{equation}
\begin{aligned}
&\mathcal{L}_{L1} = \left\| \hat{\mathbf{I}}_{t} - \mathbf{I}_{t} \right\|_{1}, \\
&\mathcal{L}_{\text{SSIM}} = 1 -\operatorname{SSIM}\left(\hat{\mathbf{I}}_{t}, \mathbf{I}_{t} \right), \\ 
&\mathcal{L}_{\text{LPIPS}} =  \left\| \phi(\hat{\mathbf{I}}_{t}) - \phi(\mathbf{I}_{t}) \right\|_{1}.
\end{aligned}
\end{equation}
The overall loss function to optimize the MPI branch would then be a sum of these three losses:
\begin{equation}
\begin{aligned}
\mathcal{L}_{\text{MPI}} = \mathcal{L}_{L1} + \mathcal{L}_{\text{SSIM}} + \mathcal{L}_{\text{LPIPS}}.
\end{aligned}
\end{equation}

These conventional loss functions train both branches to give predictions based on training views. Based on the investigation in Sec. \ref{sec:analysis}, we aim to guide the training process of the NeRF with a multiplane prior learned by the MPI branch. An intuitive choice is sampling a random number of pixels from an unseen view and matching the predicted color of two branches with an MSE loss. 
\begin{equation}
\begin{aligned}
\mathcal{L}_{\text{Mul}}= \sum 
\left\| \hat{\operatorname{C}}\left(\mathbf{r}_{i,j} \right) - \hat{\mathbf{I}}_{i,j} \right\|_{2}^{2} .
\end{aligned}
\end{equation}
$i, j$ denotes the sampled pixel coordinates in the unseen viewpoint. In practice, this intuitive choice works surprisingly well and the advantage of MPI is implicitly learned by NeRF. We give a further analysis of our design choice of $\mathcal{L}_{\text{Mul}}$ in Sec. \ref{ablation}.

To summarize, the final objective functions of the NeRF branch $\mathbf{G}_{\theta_{1}}$ and MPI branch $\mathbf{G}_{\theta_{2}}$ are given as follows:
\begin{equation}
\begin{aligned}
&\mathcal{L}_{\mathbf{G}_{\theta_{1}}}= \mathcal{L}_{\text{NeRF}} +  \lambda \mathcal{L}_{\text{Mul}},\\
&\mathcal{L}_{\mathbf{G}_{\theta_{2}}}= \mathcal{L}_{\text{MPI}}. \label{final loss}
\end{aligned}
\end{equation}
In this way, the training experience of the MPI branch serves as a multiplane prior that guides the training process of NeRF. Even if the learned MPI is not entirely accurate, the NeRF branch benefits from this multiplane prior and thus avoids collapse during training.

\section{Experiment}
\label{sec:experiment}

\begin{table*}[htbp]
  \centering
    \begin{adjustbox}{width=\linewidth}
    \begin{tabular}{c|c|ccccccccccccccccc}
    \toprule
    \toprule
    \multicolumn{1}{c}{\rotatebox{60}{Metrics}} & \multicolumn{1}{c}{\rotatebox{60}{Methods}} & \rotatebox{60}{Building} & \rotatebox{60}{Church} & \rotatebox{60}{College} & \rotatebox{60}{Mountain} & \rotatebox{60}{Mountain} & \rotatebox{60}{Observation} & \rotatebox{60}{Building} & \rotatebox{60}{Town} & \rotatebox{60}{Stadium} & \rotatebox{60}{Town} & \rotatebox{60}{Mountain} & \rotatebox{60}{Town} & \rotatebox{60}{Factory} & \rotatebox{60}{Park} & \rotatebox{60}{School} & \rotatebox{60}{Downtown} & \rotatebox{60}{Mean} \\
    \midrule
    \midrule
    \multirow{9}[0]{*}{PSNR} & NeRF\cite{mildenhall2020nerf} & \cellcolor[rgb]{ 1,  .863,  .769}13.64  & \cellcolor[rgb]{ 1,  .976,  .89}12.06  & \cellcolor[rgb]{ 1,  .976,  .89}14.45  & 18.60  & 18.40  & 14.56  & \cellcolor[rgb]{ 1,  .863,  .769}14.10  & 14.98  & 15.02  & 13.18  & 20.88  & 14.11  & 14.28  & 15.44  & \cellcolor[rgb]{ 1,  .976,  .89}14.68  & \cellcolor[rgb]{ 1,  .976,  .89}13.67  & 15.13  \\
          & Mip-NeRF\cite{barron2021mip} & 12.19  & 10.57  & 12.39  & 17.26  & 16.93  & 13.06  & 12.53  & 13.11  & 14.22  & 11.89  & 20.11  & 12.25  & 13.25  & 13.95  & 12.59  & 11.67  & 13.62  \\
          & InfoNeRF\cite{kim2022infonerf} & 13.09  & 11.53  & 14.42  & 16.43  & 16.43  & 13.68  & \cellcolor[rgb]{ 1,  .976,  .89}13.99  & 15.00  & 14.87  & 12.85  & 17.26  & 12.89  & 14.87  & 15.54  & 13.83  & 12.59  & 14.33  \\
          & DietNeRF\cite{jain2021putting} & 13.44  & \cellcolor[rgb]{ 1,  .863,  .769}12.20  & \cellcolor[rgb]{ 1,  .863,  .769}14.86  & \cellcolor[rgb]{ 1,  .976,  .89}19.34  & \cellcolor[rgb]{ 1,  .976,  .89}18.67  & \cellcolor[rgb]{ 1,  .976,  .89}15.27  & 13.73  & 15.78  & \cellcolor[rgb]{ 1,  .976,  .89}16.68  & \cellcolor[rgb]{ 1,  .976,  .89}14.21  & 20.66  & \cellcolor[rgb]{ 1,  .976,  .89}14.73  & \cellcolor[rgb]{ 1,  .976,  .89}16.73  & \cellcolor[rgb]{ 1,  .976,  .89}16.55  & \cellcolor[rgb]{ 1,  .863,  .769}14.97  & 13.61  & \cellcolor[rgb]{ 1,  .863,  .769}15.71  \\
          & PixelNeRF\cite{yu2021pixelnerf} & 6.07  & 6.26  & 7.68  & 12.09  & 10.24  & 6.77  & 6.19  & 7.35  & 5.74  & 6.02  & 12.47  & 6.89  & 4.96  & 6.53  & 4.52  & 5.88  & 7.23  \\
          & PixelNeRF ft\cite{yu2021pixelnerf} & 12.44  & 11.76  & 11.74  & 17.42  & 17.15  & 14.44  & 11.18  & \cellcolor[rgb]{ 1,  .976,  .89}15.86  & \cellcolor[rgb]{ 1,  .863,  .769}19.65  & \cellcolor[rgb]{ 1,  .863,  .769}16.09  & \cellcolor[rgb]{ 1,  .863,  .769}23.99  & \cellcolor[rgb]{ 1,  .863,  .769}15.24  & 13.02  & 15.70  & 13.86  & \cellcolor[rgb]{ 1,  .863,  .769}13.86  & 15.21  \\
          & RegNeRF\cite{niemeyer2022regnerf} & 12.07  & 10.79  & 12.60  & 16.39  & 17.36  & 13.04  & 11.92  & 12.94  & 13.49  & 11.59  & 19.37  & 12.21  & 12.66  & 13.98  & 12.71  & 11.21  & 13.40  \\
          & FreeNeRF\cite{yang2023freenerf} & \cellcolor[rgb]{ 1,  .976,  .89}13.56  & 11.01  & 13.93  & \cellcolor[rgb]{ 1,  .863,  .769}20.03  & \cellcolor[rgb]{ 1,  .863,  .769}19.74  & \cellcolor[rgb]{ 1,  .863,  .769}15.29  & 13.00  & \cellcolor[rgb]{ 1,  .863,  .769}16.29  & 15.47  & 13.21  & \cellcolor[rgb]{ 1,  .976,  .89}21.59  & 13.15  & \cellcolor[rgb]{ 1,  .863,  .769}18.91  & \cellcolor[rgb]{ 1,  .863,  .769}18.23  & 13.35  & 11.87  & \cellcolor[rgb]{ 1,  .976,  .89}15.54  \\
          & Ours  & \cellcolor[rgb]{ .98,  .706,  .694}18.81  & \cellcolor[rgb]{ .98,  .706,  .694}17.93  & \cellcolor[rgb]{ .98,  .706,  .694}20.71  & \cellcolor[rgb]{ .98,  .706,  .694}25.50  & \cellcolor[rgb]{ .98,  .706,  .694}24.92  & \cellcolor[rgb]{ .98,  .706,  .694}19.56  & \cellcolor[rgb]{ .98,  .706,  .694}18.64  & \cellcolor[rgb]{ .98,  .706,  .694}21.59  & \cellcolor[rgb]{ .98,  .706,  .694}22.08  & \cellcolor[rgb]{ .98,  .706,  .694}21.20  & \cellcolor[rgb]{ .98,  .706,  .694}28.57  & \cellcolor[rgb]{ .98,  .706,  .694}21.41  & \cellcolor[rgb]{ .98,  .706,  .694}22.61  & \cellcolor[rgb]{ .98,  .706,  .694}23.57  & \cellcolor[rgb]{ .98,  .706,  .694}20.71  & \cellcolor[rgb]{ .98,  .706,  .694}19.73  & \cellcolor[rgb]{ .98,  .706,  .694}21.72  \\
    \midrule
    \multirow{9}[0]{*}{SSIM} & NeRF\cite{mildenhall2020nerf} & \cellcolor[rgb]{ 1,  .863,  .769}0.16  & \cellcolor[rgb]{ 1,  .976,  .89}0.12  & \cellcolor[rgb]{ 1,  .976,  .89}0.17  & 0.30  & 0.24  & 0.16  & \cellcolor[rgb]{ 1,  .863,  .769}0.15  & 0.14  & 0.20  & 0.17  & 0.34  & 0.22  & 0.22  & 0.20  & \cellcolor[rgb]{ 1,  .976,  .89}0.23  & \cellcolor[rgb]{ 1,  .976,  .89}0.22  & 0.20  \\
          & Mip-NeRF\cite{barron2021mip} & 0.12  & 0.09  & 0.16  & 0.30  & 0.24  & 0.14  & 0.14  & 0.13  & 0.21  & 0.15  & 0.35  & 0.17  & 0.24  & 0.17  & 0.21  & 0.12  & 0.18  \\
          & InfoNeRF\cite{kim2022infonerf} & 0.15  & 0.12  & 0.17  & 0.26  & 0.21  & 0.14  & \cellcolor[rgb]{ 1,  .976,  .89}0.15  & 0.15  & 0.22  & 0.15  & 0.28  & 0.17  & 0.30  & 0.24  & 0.21  & 0.11  & 0.19  \\
          & DietNeRF\cite{jain2021putting} & 0.16  & \cellcolor[rgb]{ 1,  .863,  .769}0.15  & \cellcolor[rgb]{ 1,  .863,  .769}0.23  & \cellcolor[rgb]{ 1,  .976,  .89}0.34  & \cellcolor[rgb]{ 1,  .976,  .89}0.24  & \cellcolor[rgb]{ 1,  .976,  .89}0.21  & 0.17  & 0.18  & \cellcolor[rgb]{ 1,  .976,  .89}0.28  & \cellcolor[rgb]{ 1,  .976,  .89}0.21  & 0.35  & \cellcolor[rgb]{ 1,  .976,  .89}0.26  & \cellcolor[rgb]{ 1,  .976,  .89}0.36  & \cellcolor[rgb]{ 1,  .976,  .89}0.26  & \cellcolor[rgb]{ 1,  .863,  .769}0.28  & 0.19  & \cellcolor[rgb]{ 1,  .863,  .769}0.24  \\
          & PixelNeRF\cite{yu2021pixelnerf} & 0.01  & 0.01  & 0.01  & 0.02  & 0.01  & 0.01  & 0.01  & 0.01  & 0.01  & 0.01  & 0.02  & 0.01  & 0.01  & 0.00  & 0.01  & 0.01  & 0.01  \\
          & PixelNeRF ft\cite{yu2021pixelnerf} & 0.14  & 0.20  & 0.16  & 0.29  & 0.30  & 0.29  & 0.16  & \cellcolor[rgb]{ 1,  .976,  .89}0.30  & \cellcolor[rgb]{ 1,  .863,  .769}0.51  & \cellcolor[rgb]{ 1,  .863,  .769}0.47  & \cellcolor[rgb]{ 1,  .863,  .769}0.48  & \cellcolor[rgb]{ 1,  .863,  .769}0.34  & 0.25  & 0.26  & 0.29  & \cellcolor[rgb]{ 1,  .863,  .769}0.25  & 0.29  \\
          & RegNeRF\cite{niemeyer2022regnerf} & 0.12  & 0.11  & 0.16  & 0.28  & 0.27  & 0.14  & 0.13  & 0.13  & 0.20  & 0.13  & 0.34  & 0.16  & 0.20  & 0.16  & 0.21  & 0.11  & 0.18  \\
          & FreeNeRF\cite{yang2023freenerf} & \cellcolor[rgb]{ 1,  .976,  .89}0.24  & 0.11  & 0.21  & \cellcolor[rgb]{ 1,  .863,  .769}0.37  & \cellcolor[rgb]{ 1,  .863,  .769}0.33  & \cellcolor[rgb]{ 1,  .863,  .769}0.26  & 0.17  & \cellcolor[rgb]{ 1,  .863,  .769}0.28  & 0.27  & 0.23  & \cellcolor[rgb]{ 1,  .976,  .89}0.38  & 0.21  & \cellcolor[rgb]{ 1,  .863,  .769}0.51  & \cellcolor[rgb]{ 1,  .863,  .769}0.42  & 0.24  & 0.14  & \cellcolor[rgb]{ 1,  .976,  .89}0.27  \\
          & Ours  & \cellcolor[rgb]{ .98,  .706,  .694}0.73  & \cellcolor[rgb]{ .98,  .706,  .694}0.72  & \cellcolor[rgb]{ .98,  .706,  .694}0.79  & \cellcolor[rgb]{ .98,  .706,  .694}0.82  & \cellcolor[rgb]{ .98,  .706,  .694}0.81  & \cellcolor[rgb]{ .98,  .706,  .694}0.73  & \cellcolor[rgb]{ .98,  .706,  .694}0.71  & \cellcolor[rgb]{ .98,  .706,  .694}0.81  & \cellcolor[rgb]{ .98,  .706,  .694}0.80  & \cellcolor[rgb]{ .98,  .706,  .694}0.84  & \cellcolor[rgb]{ .98,  .706,  .694}0.89  & \cellcolor[rgb]{ .98,  .706,  .694}0.84  & \cellcolor[rgb]{ .98,  .706,  .694}0.86  & \cellcolor[rgb]{ .98,  .706,  .694}0.85  & \cellcolor[rgb]{ .98,  .706,  .694}0.79  & \cellcolor[rgb]{ .98,  .706,  .694}0.76  & \cellcolor[rgb]{ .98,  .706,  .694}0.80  \\
    \midrule
    \multirow{9}[0]{*}{LPIPS} & NeRF\cite{mildenhall2020nerf} & \cellcolor[rgb]{ 1,  .976,  .89}0.59  & 0.62  & \cellcolor[rgb]{ 1,  .976,  .89}0.60  & \cellcolor[rgb]{ 1,  .976,  .89}0.56  & \cellcolor[rgb]{ 1,  .976,  .89}0.58  & \cellcolor[rgb]{ 1,  .976,  .89}0.58  & \cellcolor[rgb]{ 1,  .976,  .89}0.61  & 0.59  & 0.60  & 0.59  & \cellcolor[rgb]{ 1,  .863,  .769}0.53  & \cellcolor[rgb]{ 1,  .976,  .89}0.59  & 0.53  & 0.57  & \cellcolor[rgb]{ 1,  .863,  .769}0.55  & 0.66  & \cellcolor[rgb]{ 1,  .976,  .89}0.58  \\
          & Mip-NeRF\cite{barron2021mip} & 0.64  & 0.66  & 0.66  & 0.60  & 0.64  & 0.64  & 0.62  & 0.62  & 0.61  & 0.63  & 0.60  & 0.64  & 0.56  & 0.62  & 0.63  & 0.65  & 0.63  \\
          & InfoNeRF\cite{kim2022infonerf} & 0.60  & \cellcolor[rgb]{ 1,  .863,  .769}0.60  & \cellcolor[rgb]{ 1,  .976,  .89}0.60  & 0.57  & 0.59  & 0.59  & \cellcolor[rgb]{ 1,  .976,  .89}0.61  & 0.68  & 0.58  & 0.60  & 0.57  & 0.61  & 0.53  & 0.55  & 0.60  & 0.62  & 0.59  \\
          & DietNeRF\cite{jain2021putting} & \cellcolor[rgb]{ 1,  .976,  .89}0.59  & \cellcolor[rgb]{ 1,  .976,  .89}0.61  & \cellcolor[rgb]{ 1,  .863,  .769}0.59  & \cellcolor[rgb]{ 1,  .976,  .89}0.56  & 0.59  & \cellcolor[rgb]{ 1,  .863,  .769}0.56  & \cellcolor[rgb]{ 1,  .976,  .89}0.61  & \cellcolor[rgb]{ 1,  .976,  .89}0.58  & \cellcolor[rgb]{ 1,  .976,  .89}0.52  & \cellcolor[rgb]{ 1,  .976,  .89}0.56  & \cellcolor[rgb]{ 1,  .976,  .89}0.56  & \cellcolor[rgb]{ 1,  .863,  .769}0.58  & \cellcolor[rgb]{ 1,  .976,  .89}0.48  & \cellcolor[rgb]{ 1,  .976,  .89}0.54  & \cellcolor[rgb]{ 1,  .976,  .89}0.57  & \cellcolor[rgb]{ 1,  .976,  .89}0.59  & \cellcolor[rgb]{ 1,  .863,  .769}0.57  \\
          & PixelNeRF\cite{yu2021pixelnerf} & 0.74  & 0.73  & 0.75  & 0.74  & 0.74  & 0.74  & 0.74  & 0.73  & 0.74  & 0.74  & 0.72  & 0.73  & 0.74  & 0.74  & 0.74  & 0.74  & 0.74  \\
          & PixelNeRF ft\cite{yu2021pixelnerf} & 0.70  & \cellcolor[rgb]{ 1,  .976,  .89}0.61  & 0.72  & 0.59  & 0.59  & \cellcolor[rgb]{ 1,  .976,  .89}0.58  & 0.67  & 0.59  & \cellcolor[rgb]{ 1,  .863,  .769}0.48  & \cellcolor[rgb]{ 1,  .863,  .769}0.52  & \cellcolor[rgb]{ 1,  .976,  .89}0.56  & 0.61  & 0.67  & 0.65  & 0.67  & \cellcolor[rgb]{ 1,  .863,  .769}0.58  & 0.61  \\
          & RegNeRF\cite{niemeyer2022regnerf} & 0.65  & 0.65  & 0.67  & 0.61  & 0.62  & 0.63  & 0.63  & 0.62  & 0.63  & 0.64  & 0.59  & 0.65  & 0.58  & 0.62  & 0.64  & 0.66  & 0.63  \\
          & FreeNeRF\cite{yang2023freenerf} & 0.61  & 0.67  & 0.64  & \cellcolor[rgb]{ 1,  .863,  .769}0.55  & \cellcolor[rgb]{ 1,  .976,  .89}0.58  & \cellcolor[rgb]{ 1,  .976,  .89}0.58  & 0.63  & \cellcolor[rgb]{ 1,  .976,  .89}0.58  & 0.61  & 0.61  & 0.57  & 0.64  & \cellcolor[rgb]{ 1,  .976,  .89}0.48  & \cellcolor[rgb]{ 1,  .976,  .89}0.54  & 0.61  & 0.65  & 0.60  \\
          & Ours  & \cellcolor[rgb]{ .98,  .706,  .694}0.21  & \cellcolor[rgb]{ .98,  .706,  .694}0.24  & \cellcolor[rgb]{ .98,  .706,  .694}0.18  & \cellcolor[rgb]{ .98,  .706,  .694}0.20  & \cellcolor[rgb]{ .98,  .706,  .694}0.18  & \cellcolor[rgb]{ .98,  .706,  .694}0.25  & \cellcolor[rgb]{ .98,  .706,  .694}0.24  & \cellcolor[rgb]{ .98,  .706,  .694}0.18  & \cellcolor[rgb]{ .98,  .706,  .694}0.20  & \cellcolor[rgb]{ .98,  .706,  .694}0.16  & \cellcolor[rgb]{ .98,  .706,  .694}0.12  & \cellcolor[rgb]{ .98,  .706,  .694}0.17  & \cellcolor[rgb]{ .98,  .706,  .694}0.12  & \cellcolor[rgb]{ .98,  .706,  .694}0.14  & \cellcolor[rgb]{ .98,  .706,  .694}0.20  & \cellcolor[rgb]{ .98,  .706,  .694}0.19  & \cellcolor[rgb]{ .98,  .706,  .694}0.19  \\
        \midrule
        \midrule
    \end{tabular}%
      \end{adjustbox}
    
  \caption[toc entry]{
   \textbf{Quantitative comparison with different baseline methods in 3 views.} Our MPNeRF achieves the best results compared to prior arts for few-shot neural rendering in indoor and synthetic scenes. The best, second-best, and third-best entries are marked in     
   \begin{tikzpicture}
        \draw[fill=bestcolor, draw=white] (0,0) rectangle (0.6,0.25);
    \end{tikzpicture}, 
    \begin{tikzpicture}
        \draw[fill=secondcolor, draw=white] (0,0) rectangle (0.6,0.25);
    \end{tikzpicture}, and 
    \begin{tikzpicture}
        \draw[fill=thirdcolor, draw=white] (0,0) rectangle (0.6,0.25);
    \end{tikzpicture}, 
    respectively.
  }
  \label{tab:1}
  
\end{table*}%

\begin{table*}[htbp]
  \centering
    \begin{adjustbox}{width=\linewidth}
    \begin{tabular}{c||c|ccccccccccccccccc}
    \toprule
    \toprule
    \multicolumn{1}{c}{\rotatebox{60}{Metrics}} & \multicolumn{1}{c}{\rotatebox{60}{Methods}} & \rotatebox{60}{Building} & \rotatebox{60}{Church} & \rotatebox{60}{College} & \rotatebox{60}{Mountain} & \rotatebox{60}{Mountain} & \rotatebox{60}{Observation} & \rotatebox{60}{Building} & \rotatebox{60}{Town} & \rotatebox{60}{Stadium} & \rotatebox{60}{Town} & \rotatebox{60}{Mountain} & \rotatebox{60}{Town} & \rotatebox{60}{Factory} & \rotatebox{60}{Park} & \rotatebox{60}{School} & \rotatebox{60}{Downtown} & \rotatebox{60}{Mean} \\
    \midrule
    \midrule
    \multicolumn{1}{c|}{\multirow{9}[2]{*}{PSNR}} & NeRF\cite{mildenhall2020nerf} & 13.44  & 12.64  & 15.62  & 19.91  & 19.39  & 15.79  & 15.11  & 16.69  & 16.44  & 14.90  & 22.06  & 13.44  & 13.76  & 15.25  & 13.67  & 13.31  & 15.71  \\
    \multicolumn{1}{c|}{} & Mip-NeRF\cite{barron2021mip} & 12.42  & 10.94  & 12.91  & 18.13  & 17.24  & 14.14  & 12.74  & 13.65  & 15.08  & 12.40  & 20.79  & 13.21  & 14.33  & 14.78  & 13.08  & 11.94  & 14.24  \\
    \multicolumn{1}{c|}{} & InfoNeRF \cite{kim2022infonerf} & 13.31  & 12.30  & 15.32  & 18.82  & 18.63  & 15.90  & 14.77  & 15.94  & 15.92  & 13.20  & 21.57  & 15.17  & 16.05  & 15.90  & 14.87  & 13.22  & 15.68  \\
    \multicolumn{1}{c|}{} & DietNeRF \cite{jain2021putting} & 13.82  & 13.01  & \cellcolor[rgb]{ 1,  .976,  .89}16.35  & 20.35  & 19.67  & 16.13  & \cellcolor[rgb]{ 1,  .976,  .89}15.45  & \cellcolor[rgb]{ 1,  .976,  .89}16.84  & \cellcolor[rgb]{ 1,  .863,  .769}17.31  & 15.03  & 22.49  & \cellcolor[rgb]{ 1,  .976,  .89}16.30  & \cellcolor[rgb]{ 1,  .976,  .89}17.86  & \cellcolor[rgb]{ 1,  .976,  .89}17.59  & \cellcolor[rgb]{ 1,  .976,  .89}15.66  & 14.77  & 16.79  \\
    \multicolumn{1}{c|}{} & PixelNeRF \cite{yu2021pixelnerf} & 6.07  & 6.31  & 7.81  & 12.03  & 10.22  & 6.84  & 6.29  & 7.41  & 5.80  & 6.03  & 12.45  & 6.85  & 5.04  & 6.59  & 4.58  & 5.92  & 7.27  \\
    \multicolumn{1}{c|}{} & PixelNeRF ft \cite{yu2021pixelnerf} & \cellcolor[rgb]{ 1,  .976,  .89}15.67  & \cellcolor[rgb]{ 1,  .976,  .89}15.05  & 15.84  & 21.17  & \cellcolor[rgb]{ 1,  .976,  .89}21.01  & 16.26  & 15.27  & 16.58  & \cellcolor[rgb]{ 1,  .976,  .89}17.18  & \cellcolor[rgb]{ 1,  .976,  .89}15.21  & \cellcolor[rgb]{ 1,  .976,  .89}22.65  & 16.03  & 14.90  & 16.52  & 15.36  & \cellcolor[rgb]{ 1,  .976,  .89}14.83  & \cellcolor[rgb]{ 1,  .976,  .89}16.85  \\
    \multicolumn{1}{c|}{} & RegNeRF\cite{niemeyer2022regnerf} & 12.20  & 12.57  & 14.09  & \cellcolor[rgb]{ 1,  .863,  .769}22.16  & 19.00  & \cellcolor[rgb]{ 1,  .976,  .89}17.02  & 13.79  & 14.07  & 14.16  & 12.48  & 20.67  & 12.76  & 16.23  & 14.47  & 12.94  & 11.82  & 15.03  \\
    \multicolumn{1}{c|}{} & FreeNeRF\cite{yang2023freenerf} & \cellcolor[rgb]{ 1,  .863,  .769}16.83  & \cellcolor[rgb]{ 1,  .863,  .769}16.66  & \cellcolor[rgb]{ 1,  .863,  .769}19.54  & \cellcolor[rgb]{ 1,  .976,  .89}21.97  & \cellcolor[rgb]{ 1,  .863,  .769}21.44  & \cellcolor[rgb]{ 1,  .863,  .769}18.28  & \cellcolor[rgb]{ 1,  .863,  .769}17.93  & \cellcolor[rgb]{ 1,  .863,  .769}19.81  & 16.63  & \cellcolor[rgb]{ 1,  .863,  .769}17.90  & \cellcolor[rgb]{ 1,  .863,  .769}23.95  & \cellcolor[rgb]{ 1,  .863,  .769}20.37  & \cellcolor[rgb]{ 1,  .863,  .769}21.42  & \cellcolor[rgb]{ 1,  .863,  .769}17.90  & \cellcolor[rgb]{ 1,  .863,  .769}18.25  & \cellcolor[rgb]{ 1,  .863,  .769}15.77  & \cellcolor[rgb]{ 1,  .863,  .769}19.04  \\
    \multicolumn{1}{c|}{} & Ours  & \cellcolor[rgb]{ .98,  .706,  .694}20.50  & \cellcolor[rgb]{ .98,  .706,  .694}19.56  & \cellcolor[rgb]{ .98,  .706,  .694}23.08  & \cellcolor[rgb]{ .98,  .706,  .694}26.02  & \cellcolor[rgb]{ .98,  .706,  .694}24.88  & \cellcolor[rgb]{ .98,  .706,  .694}21.29  & \cellcolor[rgb]{ .98,  .706,  .694}20.99  & \cellcolor[rgb]{ .98,  .706,  .694}21.92  & \cellcolor[rgb]{ .98,  .706,  .694}23.07  & \cellcolor[rgb]{ .98,  .706,  .694}21.57  & \cellcolor[rgb]{ .98,  .706,  .694}29.00  & \cellcolor[rgb]{ .98,  .706,  .694}22.19  & \cellcolor[rgb]{ .98,  .706,  .694}22.58  & \cellcolor[rgb]{ .98,  .706,  .694}23.59  & \cellcolor[rgb]{ .98,  .706,  .694}21.72  & \cellcolor[rgb]{ .98,  .706,  .694}20.50  & \cellcolor[rgb]{ .98,  .706,  .694}22.65  \\
    \midrule
    \multicolumn{1}{c|}{\multirow{9}[0]{*}{SSIM}} & NeRF\cite{mildenhall2020nerf} & \cellcolor[rgb]{ 1,  .976,  .89}0.16  & 0.12  & 0.17  & 0.30  & 0.24  & 0.16  & 0.15  & 0.14  & 0.20  & 0.17  & 0.34  & 0.22  & 0.22  & 0.20  & 0.23  & 0.22  & 0.20  \\
    \multicolumn{1}{c|}{} & Mip-NeRF\cite{barron2021mip} & 0.12  & 0.09  & 0.15  & 0.30  & 0.23  & 0.17  & 0.14  & 0.13  & 0.21  & 0.14  & 0.36  & 0.17  & 0.25  & 0.18  & 0.21  & 0.12  & 0.19  \\
    \multicolumn{1}{c|}{} & InfoNeRF\cite{kim2022infonerf} & 0.15  & 0.12  & 0.17  & 0.26  & 0.21  & 0.14  & 0.15  & 0.15  & 0.22  & 0.15  & 0.28  & 0.17  & 0.30  & 0.24  & 0.21  & 0.11  & 0.19  \\
    \multicolumn{1}{c|}{} & DietNeRF\cite{jain2021putting} & \cellcolor[rgb]{ 1,  .976,  .89}0.16  & 0.15  & 0.23  & 0.34  & 0.24  & 0.21  & 0.17  & 0.18  & 0.28  & 0.21  & 0.35  & 0.26  & 0.36  & \cellcolor[rgb]{ 1,  .976,  .89}0.26  & 0.28  & 0.19  & 0.24  \\
    \multicolumn{1}{c|}{} & PixelNeRF\cite{yu2021pixelnerf} & 0.01  & 0.01  & 0.01  & 0.02  & 0.01  & 0.01  & 0.01  & 0.01  & 0.01  & 0.01  & 0.02  & 0.01  & 0.01  & 0.00  & 0.01  & 0.01  & 0.01  \\
    \multicolumn{1}{c|}{} & PixelNeRF(ft)\cite{yu2021pixelnerf} & 0.14  & 0.20  & 0.16  & 0.29  & 0.30  & 0.29  & 0.16  & \cellcolor[rgb]{ 1,  .976,  .89}0.30  & \cellcolor[rgb]{ 1,  .863,  .769}0.51  & \cellcolor[rgb]{ 1,  .976,  .89}0.47  & \cellcolor[rgb]{ 1,  .863,  .769}0.48  & \cellcolor[rgb]{ 1,  .976,  .89}0.34  & 0.25  & \cellcolor[rgb]{ 1,  .976,  .89}0.26  & \cellcolor[rgb]{ 1,  .976,  .89}0.29  & \cellcolor[rgb]{ 1,  .976,  .89}0.25  & \cellcolor[rgb]{ 1,  .976,  .89}0.29  \\
    \multicolumn{1}{c|}{} & RegNeRF\cite{niemeyer2022regnerf} & 0.12  & \cellcolor[rgb]{ 1,  .976,  .89}0.24  & \cellcolor[rgb]{ 1,  .976,  .89}0.26  & \cellcolor[rgb]{ 1,  .863,  .769}0.54  & \cellcolor[rgb]{ 1,  .976,  .89}0.36  & \cellcolor[rgb]{ 1,  .863,  .769}0.40  & \cellcolor[rgb]{ 1,  .976,  .89}0.25  & 0.17  & 0.20  & 0.15  & 0.35  & 0.17  & \cellcolor[rgb]{ 1,  .976,  .89}0.44  & 0.17  & 0.20  & 0.11  & 0.26  \\
    \multicolumn{1}{c|}{} & FreeNeRF\cite{yang2023freenerf} & \cellcolor[rgb]{ 1,  .863,  .769}0.47  & \cellcolor[rgb]{ 1,  .863,  .769}0.44  & \cellcolor[rgb]{ 1,  .863,  .769}0.44  & \cellcolor[rgb]{ 1,  .976,  .89}0.44  & \cellcolor[rgb]{ 1,  .863,  .769}0.39  & \cellcolor[rgb]{ 1,  .976,  .89}0.35  & \cellcolor[rgb]{ 1,  .863,  .769}0.41  & \cellcolor[rgb]{ 1,  .863,  .769}0.47  & \cellcolor[rgb]{ 1,  .976,  .89}0.33  & \cellcolor[rgb]{ 1,  .863,  .769}0.48  & \cellcolor[rgb]{ 1,  .976,  .89}0.44  & \cellcolor[rgb]{ 1,  .863,  .769}0.54  & \cellcolor[rgb]{ 1,  .863,  .769}0.65  & \cellcolor[rgb]{ 1,  .863,  .769}0.41  & \cellcolor[rgb]{ 1,  .863,  .769}0.46  & \cellcolor[rgb]{ 1,  .863,  .769}0.37  & \cellcolor[rgb]{ 1,  .863,  .769}0.44  \\
    \multicolumn{1}{c|}{} & Ours  & \cellcolor[rgb]{ .98,  .706,  .694}0.81  & \cellcolor[rgb]{ .98,  .706,  .694}0.80  & \cellcolor[rgb]{ .98,  .706,  .694}0.87  & \cellcolor[rgb]{ .98,  .706,  .694}0.87  & \cellcolor[rgb]{ .98,  .706,  .694}0.83  & \cellcolor[rgb]{ .98,  .706,  .694}0.81  & \cellcolor[rgb]{ .98,  .706,  .694}0.81  & \cellcolor[rgb]{ .98,  .706,  .694}0.84  & \cellcolor[rgb]{ .98,  .706,  .694}0.85  & \cellcolor[rgb]{ .98,  .706,  .694}0.87  & \cellcolor[rgb]{ .98,  .706,  .694}0.90  & \cellcolor[rgb]{ .98,  .706,  .694}0.87  & \cellcolor[rgb]{ .98,  .706,  .694}0.87  & \cellcolor[rgb]{ .98,  .706,  .694}0.86  & \cellcolor[rgb]{ .98,  .706,  .694}0.83  & \cellcolor[rgb]{ .98,  .706,  .694}0.81  & \cellcolor[rgb]{ .98,  .706,  .694}0.84  \\
    \midrule
    \multicolumn{1}{c|}{\multirow{9}[0]{*}{LPIPS}} & NeRF\cite{mildenhall2020nerf} & \cellcolor[rgb]{ 1,  .976,  .89}0.59  & 0.62  & \cellcolor[rgb]{ 1,  .976,  .89}0.60  & \cellcolor[rgb]{ 1,  .976,  .89}0.56  & \cellcolor[rgb]{ 1,  .976,  .89}0.58  & \cellcolor[rgb]{ 1,  .976,  .89}0.58  & \cellcolor[rgb]{ 1,  .976,  .89}0.61  & 0.59  & 0.60  & 0.59  & \cellcolor[rgb]{ 1,  .863,  .769}0.53  & \cellcolor[rgb]{ 1,  .976,  .89}0.59  & 0.53  & 0.57  & \cellcolor[rgb]{ 1,  .976,  .89}0.55  & 0.66  & \cellcolor[rgb]{ 1,  .976,  .89}0.58  \\
    \multicolumn{1}{c|}{} & InfoNeRF\cite{kim2022infonerf} & 0.60  & \cellcolor[rgb]{ 1,  .863,  .769}0.60  & \cellcolor[rgb]{ 1,  .976,  .89}0.60  & 0.57  & 0.59  & 0.59  & \cellcolor[rgb]{ 1,  .976,  .89}0.61  & 0.68  & 0.58  & 0.60  & 0.57  & 0.61  & 0.53  & \cellcolor[rgb]{ 1,  .976,  .89}0.55  & 0.60  & 0.62  & 0.59  \\
    \multicolumn{1}{c|}{} & Mip-NeRF\cite{barron2021mip} & 0.63  & 0.65  & 0.65  & 0.59  & 0.63  & 0.60  & 0.62  & 0.61  & 0.60  & 0.62  & 0.58  & 0.63  & 0.55  & 0.60  & 0.62  & 0.64  & 0.61  \\
    \multicolumn{1}{c|}{} & DietNeRF\cite{jain2021putting} & \cellcolor[rgb]{ 1,  .976,  .89}0.59  & \cellcolor[rgb]{ 1,  .976,  .89}0.61  & \cellcolor[rgb]{ 1,  .863,  .769}0.59  & \cellcolor[rgb]{ 1,  .976,  .89}0.56  & 0.59  & \cellcolor[rgb]{ 1,  .863,  .769}0.56  & \cellcolor[rgb]{ 1,  .976,  .89}0.61  & \cellcolor[rgb]{ 1,  .976,  .89}0.58  & \cellcolor[rgb]{ 1,  .976,  .89}0.52  & \cellcolor[rgb]{ 1,  .976,  .89}0.56  & \cellcolor[rgb]{ 1,  .976,  .89}0.56  & \cellcolor[rgb]{ 1,  .976,  .89}0.58  & \cellcolor[rgb]{ 1,  .976,  .89}0.48  & \cellcolor[rgb]{ 1,  .863,  .769}0.54  & \cellcolor[rgb]{ 1,  .976,  .89}0.57  & \cellcolor[rgb]{ 1,  .976,  .89}0.59  & \cellcolor[rgb]{ 1,  .976,  .89}0.57  \\
    \multicolumn{1}{c|}{} & PixelNeRF\cite{yu2021pixelnerf} & 0.74  & 0.73  & 0.75  & 0.74  & 0.74  & 0.74  & 0.74  & 0.73  & 0.74  & 0.74  & 0.72  & 0.73  & 0.74  & 0.74  & 0.74  & 0.74  & 0.74  \\
    \multicolumn{1}{c|}{} & PixelNeRF(ft)\cite{yu2021pixelnerf} & 0.70  & \cellcolor[rgb]{ 1,  .976,  .89}0.61  & 0.72  & 0.59  & 0.59  & \cellcolor[rgb]{ 1,  .976,  .89}0.58  & 0.67  & 0.59  & \cellcolor[rgb]{ 1,  .863,  .769}0.48  & \cellcolor[rgb]{ 1,  .976,  .89}0.52  & \cellcolor[rgb]{ 1,  .976,  .89}0.56  & 0.61  & 0.67  & 0.65  & 0.67  & \cellcolor[rgb]{ 1,  .976,  .89}0.58  & 0.61  \\
    \multicolumn{1}{c|}{} & RegNeRF\cite{niemeyer2022regnerf} & 0.64  & \cellcolor[rgb]{ 1,  .976,  .89}0.56  & \cellcolor[rgb]{ 1,  .976,  .89}0.58  & \cellcolor[rgb]{ 1,  .863,  .769}0.43  & \cellcolor[rgb]{ 1,  .976,  .89}0.55  & \cellcolor[rgb]{ 1,  .863,  .769}0.47  & \cellcolor[rgb]{ 1,  .976,  .89}0.58  & \cellcolor[rgb]{ 1,  .976,  .89}0.58  & 0.62  & 0.61  & 0.57  & 0.63  & \cellcolor[rgb]{ 1,  .976,  .89}0.46  & 0.60  & 0.62  & 0.65  & 0.57  \\
    \multicolumn{1}{c|}{} & FreeNeRF\cite{yang2023freenerf} & \cellcolor[rgb]{ 1,  .863,  .769}0.52  & \cellcolor[rgb]{ 1,  .863,  .769}0.50  & \cellcolor[rgb]{ 1,  .863,  .769}0.52  & \cellcolor[rgb]{ 1,  .976,  .89}0.52  & \cellcolor[rgb]{ 1,  .976,  .89}0.55  & \cellcolor[rgb]{ 1,  .976,  .89}0.53  & \cellcolor[rgb]{ 1,  .863,  .769}0.53  & \cellcolor[rgb]{ 1,  .863,  .769}0.50  & 0.56  & \cellcolor[rgb]{ 1,  .863,  .769}0.48  & \cellcolor[rgb]{ 1,  .976,  .89}0.55  & \cellcolor[rgb]{ 1,  .863,  .769}0.47  & \cellcolor[rgb]{ 1,  .863,  .769}0.43  & \cellcolor[rgb]{ 1,  .976,  .89}0.55  & \cellcolor[rgb]{ 1,  .863,  .769}0.51  & \cellcolor[rgb]{ 1,  .863,  .769}0.56  & \cellcolor[rgb]{ 1,  .863,  .769}0.52  \\
      \multicolumn{1}{c|}{}   & Ours  & \cellcolor[rgb]{ .98,  .706,  .694}0.17  & \cellcolor[rgb]{ .98,  .706,  .694}0.20  & \cellcolor[rgb]{ .98,  .706,  .694}0.14  & \cellcolor[rgb]{ .98,  .706,  .694}0.18  & \cellcolor[rgb]{ .98,  .706,  .694}0.18  & \cellcolor[rgb]{ .98,  .706,  .694}0.21  & \cellcolor[rgb]{ .98,  .706,  .694}0.19  & \cellcolor[rgb]{ .98,  .706,  .694}0.16  & \cellcolor[rgb]{ .98,  .706,  .694}0.17  & \cellcolor[rgb]{ .98,  .706,  .694}0.15  & \cellcolor[rgb]{ .98,  .706,  .694}0.12  & \cellcolor[rgb]{ .98,  .706,  .694}0.15  & \cellcolor[rgb]{ .98,  .706,  .694}0.10  & \cellcolor[rgb]{ .98,  .706,  .694}0.12  & \cellcolor[rgb]{ .98,  .706,  .694}0.17  & \cellcolor[rgb]{ .98,  .706,  .694}0.14  & \cellcolor[rgb]{ .98,  .706,  .694}0.16  \\
    \bottomrule
    \bottomrule
    \end{tabular}%
      \end{adjustbox}
  \caption[toc entry]{
   \textbf{Quantitative comparison with different baseline methods in 5 views.} Our MPNeRF achieves the best results compared to prior arts for few-shot neural rendering in indoor and synthetic scenes. The best, second-best, and third-best entries are marked in     
   \begin{tikzpicture}
        \draw[fill=bestcolor, draw=white] (0,0) rectangle (0.6,0.25);
    \end{tikzpicture}, 
    \begin{tikzpicture}
        \draw[fill=secondcolor, draw=white] (0,0) rectangle (0.6,0.25);
    \end{tikzpicture}, and 
    \begin{tikzpicture}
        \draw[fill=thirdcolor, draw=white] (0,0) rectangle (0.6,0.25);
    \end{tikzpicture}, 
    respectively.
  }
  \label{tab:2}
  
\end{table*}%

\begin{figure*}
    \centering
    \includegraphics[width=0.9\linewidth]{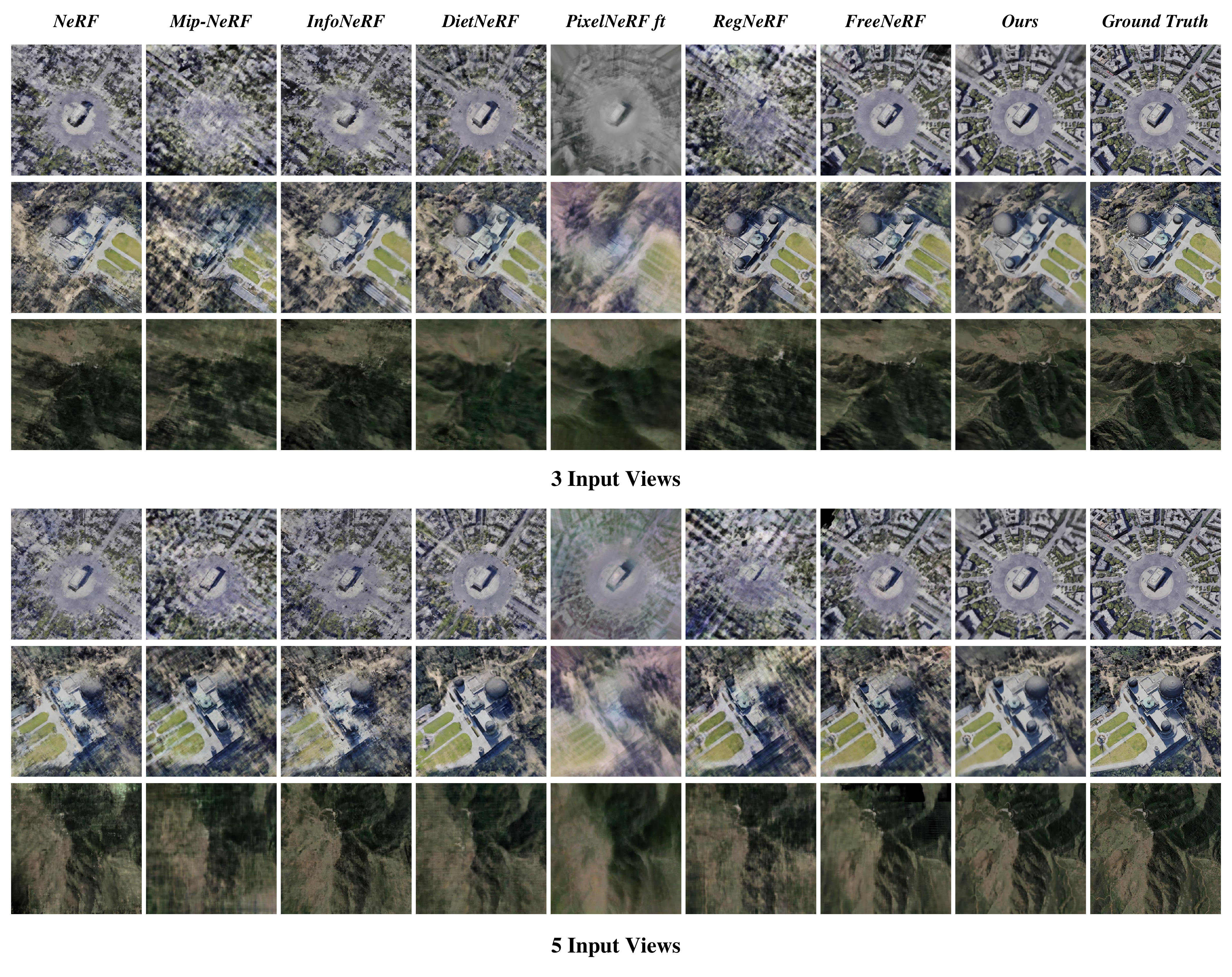}
    \caption{Visual comparisons on 3 selected scenes with 3 and 5
    views. MPNeRF achieves photo-realistic quality in different scenes compared with ground-truth images on novel views.}
    \label{fig:vis}
\end{figure*}

\subsection{Implementation Details}
Our method is implemented using PyTorch, and all experiments are conducted on a GeForce RTX 3090 GPU. 
For the NeRF branch, we use the original NeRF in \cite{mildenhall2020nerf}. During training, we randomly sample unseen views following the strategy proposed by \cite{jain2021putting}. The batch size is set to 1024 pixel rays in both source and unseen views. For each ray, we perform 64 coarse sampling and 32 fine sampling along the ray. For the MPI branch, we sample 16 layers of planes for each viewpoint. The optimization of the two branches of MPNeRF is performed using the Adam optimizer \cite{kingma2014adam} with a learning rate of $5 \times 10^{-4}$. The hyperparameter $\lambda$ in Eq. \ref{final loss} is set to $1$.

\subsection{Datasets and Evaluation Metrics}
The main experiments are conducted on 16 scenes collected by LEVIR-NVS \cite{wu2022remote}. These scenes contain various scenarios in common aerial imagery, including mountains, buildings, colleges, etc. 3 and 5 views are used for training and the rest for testing. Additional experiments and discussions can be found in the Appendix. In line with previous studies of few-shot neural rendering \cite{kim2022infonerf,jain2021putting,niemeyer2022regnerf,yang2023freenerf}, we report PSNR, SSIM and LPIPS \cite{zhang2018unreasonable}. 

\subsection{Baseline Methods}
We compare MPNeRF against various state-of-the-art methods including NeRF \cite{mildenhall2020nerf}, Mip-NeRF \cite{barron2021mip}, InfoNeRF \cite{kim2022infonerf}, DietNeRF \cite{jain2021putting}, PixelNeRF \cite{yu2021pixelnerf}, RegNeRF \cite{niemeyer2022regnerf} and FreeNeRF \cite{yang2023freenerf}. Among these methods, NeRF and Mip-NeRF are designed for dense view training, we mainly explore the performance gain achieved by MPNeRF. PixelNeRF aims to learn a generalized NeRF representation for all scenes and is pre-trained on the DTU dataset \cite{jensen2014large}. Since a large domain gap might exist when applied in aerial imagery, we report PixelNeRF's results with and without additional fine-tuning per scene. Other methods are designed for few-shot neural rendering, and we conduct comparative experiments to investigate their performance when encountering aerial imagery.

\subsection{Comparative Results Analysis}
Table \ref{tab:1} and Table \ref{tab:2} report the performance of MPNeRF and baseline methods in the 3-view and 5-view settings. Additionally, a qualitative comparison can be observed in Fig. \ref{fig:vis}. A very significant improvement can be found in all three metrics and rendering fidelity. 
The results demonstrate that PixelNeRF tends to produce blurry renderings, which we attribute to the poor localization of the CNN features. InfoNeRF and RegNeRF use local smoothness and sparsity to regularize NeRF explicitly. However, in scenarios with substantially limited information compared to the scene complexity, the performance of these methods could be compromised. DietNeRF implicitly distills the prior knowledge encoded in CLIP \cite{AlecRadford2021LearningTV} and achieves better results. FreeNeRF investigates the frequency in NeRF training. By progressively learning each frequency component, FreeNeRF has demonstrated remarkable effectiveness. Nonetheless, the progressive frequency regularization leads to relatively flat results, favoring PSNR but not metrics that consider local structures such as SSIM and LPIPS. 

In fact, NeRF's representation makes recovering 3D scenes from sparse inputs ill-posed. MPNeRF acquires superior results by the guiding of a multiplane prior to gaining a stronger understanding of local structures and semantics. In the more challenging scenes, such as Building in Figure \ref{fig:vis}, MPNeRF successfully avoids collapse during training.

\subsection{Ablation Studies and Further Analyses}
\label{ablation}

\noindent \textbf{Ablation Analysis.} We ablate the proposed multiplane prior to our method, and the results are shown in Table \ref{tab:abl}. 
Intuitively, it seems better to use MPI as a guide after fully training it, we first construct experiments where a two-stage training strategy is employed. We then assess the impact of SwinV2's pre-trained weights on performance by removing them. Next, we evaluate the contribution of multi-scale features by disconnecting the skip connections in the MPI generator. Finally, we integrate the multiplane directly within NeRF’s sampling space, omitting the separate MPI branch, to examine the inductive biases' influence on performance. 
Employing MPI concurrent branch during training leads to slight improvements. We believe this is because the MPI's training experience itself carries information.
The exclusion of SwinV2's pre-trained weights declines performance, affirming the value of the encoded prior knowledge. Similarly, omitting the multi-scale feature connection diminishes the fidelity of the rendered images. Most significantly, the absence of the MPI generator results in a marked decrease in all metrics. This suggests that uncalibrated MPI generated by self-attention and convolution is important to avoid degenerate solutions. Collectively, these findings demonstrate that each element of the proposed multiplane prior is crucial for the superior performance of MPNeRF.

\begin{table}[H]
\centering
\begin{tabular}{lccc}
\hline
\textbf{Methods} & \textbf{PSNR} & \textbf{SSIM} & \textbf{LPIPS} \\
\hline
Ours & \cellcolor[rgb]{ .98,  .706,  .694}21.72 & \cellcolor[rgb]{ .98,  .706,  .694}0.80 & \cellcolor[rgb]{ .98,  .706,  .694}0.19  \\
training MPI beforehand & 21.49 & 0.76 & 0.20 \\
w/o pre-trained weights & 20.46 & 0.75 & 0.24 \\
w/o multi-scale feature & 20.05 & 0.71 & 0.25 \\
w/o MPI generator & 16.38 & 0.42 & 0.52  \\
Baseline NeRF & 15.13 & 0.20 & 0.58  \\
\hline
\end{tabular}
\caption{Ablation analysis on the proposed Multiplane Prior.}
\label{tab:abl}
\end{table}

\noindent \textbf{Further Analyses on the design choice of the $\mathcal{L}_{Mul}$.} One intuitive thought of designing $\mathcal{L}_{Mul}$ is that geometry recovered by the MPI branch may provide more information than color alone. So we design two experiments, one is to match the expected depth of both branches as an auxiliary depth loss, and another is to model density on each ray as a distribution \cite{kim2022infonerf} and minimize the KL divergence.
Another intuitive thought is that the choice of $\mathcal{L}_{Mul}$ should reflect the local or nonlocal relationships within the pixels. Therefore, we adopt the recently proposed S3IM loss \cite{xie2023s3im} to measure this relationship. 
\begin{table}[H]
\centering
\begin{tabular}{lccc}
\hline
\textbf{Design Choice} & \textbf{PSNR} & \textbf{SSIM} & \textbf{LPIPS} \\
\hline
w/t depth matching & 21.34 & 0.75 & 0.20 \\
w/t ray matching   & 21.11 & 0.71 & 0.21 \\
w/t relation matching & 21.19 & 0.70 & 0.21  \\
Ours & \cellcolor[rgb]{ .98,  .706,  .694}21.72 & \cellcolor[rgb]{ .98,  .706,  .694}0.80 & \cellcolor[rgb]{ .98,  .706,  .694}0.19  \\
\hline
\end{tabular}
\caption{Design choice of the $\mathcal{L}_{Mul}$}
\label{tab:design}
\end{table}
However, as shown in Table. \ref{tab:design}, the result suggests that these intuitive designs worsen the results. The first two design involves direct supervision of the depth generated by the MPI generator. The last involves capturing the non-local relationships between the predictions of the NeRF and the MPI branch. Since the learned MPI is not entirely accurate, we believe the noise within pseudo-labels may compromise performance with these enhanced supervisions applied. 

\noindent \textbf{Impact of Different Pre-trained Models.}
We perform a comparison study on three pre-trained vision transformers, i.e., CLIP \cite{AlecRadford2021LearningTV}, DINOV2 \cite{oquab2023dinov2}, and SimMIM \cite{xie2022simmim}. We adopt the base model in our experiment. As shown in Table \ref{tab:pretrain}, all of these methods provide comparable results. The results show that the Swin Transformer pre-trained via SimMIM \cite{xie2022simmim} outperforms others. We believe this can be attributed to the rich global and local details learned by SimMIM and the hierarchical structure of the Swin Transformer.

\begin{table}[H]
\centering
\begin{tabular}{lccc}
\hline
\textbf{Pre-trained Image Encoder} & \textbf{PSNR} & \textbf{SSIM} & \textbf{LPIPS} \\
\hline
CLIP \cite{AlecRadford2021LearningTV} & 20.15 & 0.70 & 0.25 \\
DINOV2 \cite{oquab2023dinov2} & 20.01 & 0.68 & 0.25 \\
SimMIM (Ours) \cite{xie2022simmim} & \cellcolor[rgb]{ .98,  .706,  .694}21.72 & \cellcolor[rgb]{ .98,  .706,  .694}0.80 & \cellcolor[rgb]{ .98,  .706,  .694}0.19  \\
\hline
\end{tabular}
\caption{Impact of different pre-trained models.}
\label{tab:pretrain}
\end{table}

\begin{figure}
    \centering
    \includegraphics[width=1\linewidth]{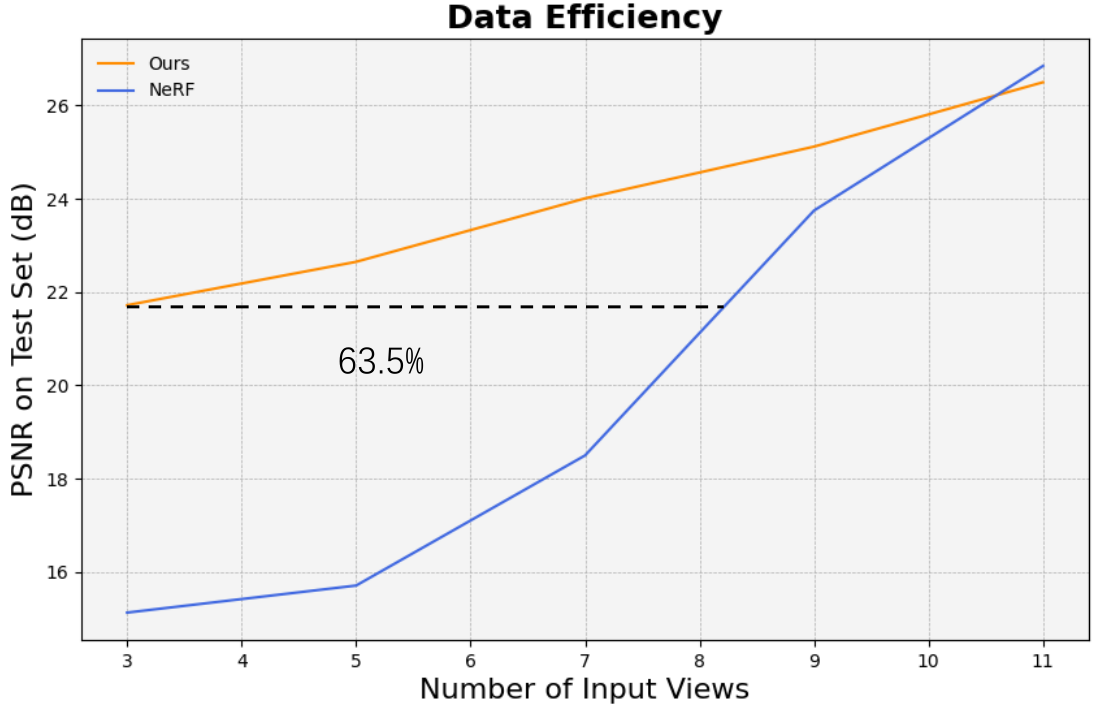}
    \caption{We investigate the data efficiency achieved by our method. Our method requires up to $63.5\%$ training images to achieve a similar performance compared to a vanilla NeRF model.}
    \label{fig:data}
\end{figure}

% \textbf{Data Efficiency.} 
\noindent \textbf{Data Efficiency}
Since we aim to improve the capability of NeRF in aerial scenes when only sparse views are available, we investigate how much data MPNeRF can save to achieve similar rendering results compared to the original NeRF that requires dense view supervision. As shown in Fig. \ref{fig:data}, the results show that our method requires up to $63.5\%$ training images. This may help save energy and establish resource-efficient applications for UAVs based on NeRF.

\section{Limitations and Conclusion}
\label{sec:conclusion}
In this work, we introduce Multiplane Prior guided NeRF (MPNeRF), the first approach designed for few-shot aerial scene rendering.
Through the guiding of the multiplane prior, MPNeRF effectively overcomes the typical pitfalls in spare aerial scenes. We hope our work can provide insight into future NeRF-based applications in aerial scenes.
However, further exploration of the guiding strategy design is needed. In particular, incorporating uncertainty prediction mechanisms or implementing grid-based representations holds promise for future research directions.

{\small
\bibliographystyle{ieeenat_fullname}
\bibliography{11_references}
}

\ifarxiv \clearpage \appendix In this supplement, we first conduct more experimental results and discussion to evaluate the robustness and efficiency of our proposed Multiplane Prior guided NeRF (MPNeRF). We also include more qualitative results to discuss the motivation and limitations of MPNeRF. Finally, we add more details of experimental settings and implementations.

\section{Additional Experiments and Analysis}
\noindent \textbf{Robustness to Hyperparameters.}
We have conducted a series of experiments to assess the sensitivity of our model to hyperparameters. Specifically, we focus on the hyperparameter $\lambda$, which plays a crucial role in balancing different components of our loss function. In Figure \ref{fig:hyp}, we illustrate the impact of varying $\lambda$ on the performance of the proposed MPNeRF and a standard NeRF \cite{mildenhall2020nerf} model. 

As $\lambda$ increases, we observe that the PSNR and SSIM metrics tend to plateau, suggesting that there is an optimal range for $\lambda$ wherein the model achieves a balance between fidelity and perceptual quality. On the other hand, the LPIPS metric shows an initial decrease followed by a gradual increase, indicating a sweet spot where the model best captures the perceptual features of the aerial scenes. The trends exhibited by MPNeRF show its relative insensitivity to $\lambda$ within a reasonable range, which underscores the robustness of our method. Notably, MPNeRF consistently outperforms the baseline NeRF model across all metrics, demonstrating the effectiveness of incorporating the multiplane prior to the rendering process.

\begin{figure}[H]
    \centering
    \includegraphics[width=1\linewidth]{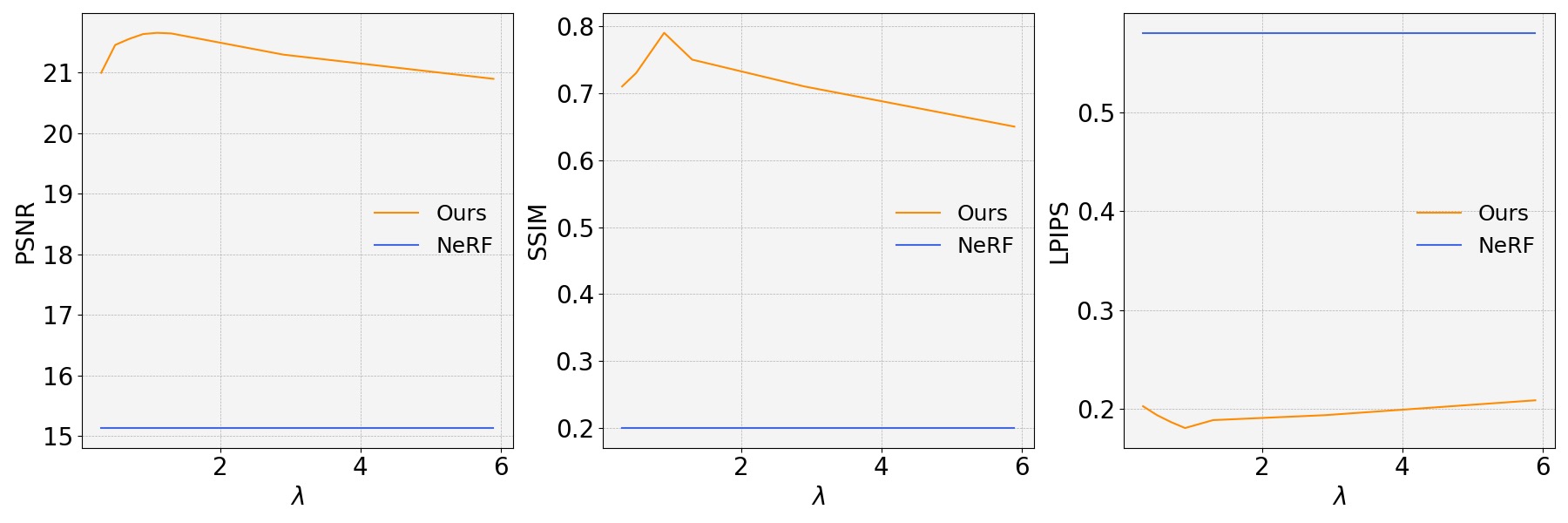}
    \caption{\textbf{Hyperparameter Sensitivity Analysis.} Performance comparison of our method (MPNeRF) and a baseline NeRF model across different values of hyperparameter $\lambda$. The graphs show PSNR, SSIM, and LPIPS metrics. MPNeRF is robust to a wide $\lambda$ choice.}
    \label{fig:hyp}
\end{figure}

\noindent \textbf{MPNeRF vs MPI-based Methods.}
The present study of MPI mainly focuses on overcoming shortcomings such as ``failure to represent continuous 3D space" in \cite{li2021mine}. In contrast, our approach utilizes MPI as a bridge to convey complex information that a single NeRF struggles with. We construct comparison experiments under 3-view settings among the proposed MPNeRF, the original MPI \cite{tucker2020single}, and MINE \cite{li2021mine}
In Table. \ref{tab:branch}, the original MPI achieves an 18.57 PSNR, 0.54 SSIM, and 0.45 LPIPS. While MINE performs better with 19.99 PSNR, 0.61 SSIM, and 0.40 LPIPS. Our MPNeRF outperforms these methods by a large margin. These MPI-based methods face inherent limitations like ghosting effects and cropped corners under sparse inputs and large camera movements.

\begin{table}[H]
\centering
\begin{tabular}{lccc}
\toprule
\textbf{Methods} & \textbf{PSNR} & \textbf{SSIM} & \textbf{LPIPS} \\
\midrule
MPI \cite{tucker2020single} & 18.57 & 0.54 & 0.45 \\
MINE \cite{li2021mine} & 19.99 & 0.61 & 0.40 \\
\midrule
NeRF branch w/o $\mathcal{L}_{mul}$  & 15.13 & 0.20 & 0.58 \\
MPI branch & 20.32 & 0.57 & 0.34 \\
\midrule
NeRF branch w/t $\mathcal{L}_{mul}$ & 21.72 & 0.80 & 0.19 \\
\bottomrule
\end{tabular}
\caption{\textbf{Comparation between the NeRF and MPI.} This table presents the evaluation of the MPI based method in previous studies, NeRF branch without multiplane loss ($\mathcal{L}_{mul}$), the MPI branch independently, and the NeRF branch with $\mathcal{L}_{mul}$ within our MPNeRF framework. The metrics of PSNR, SSIM, and LPIPS demonstrate the significant impact of the multiplane prior on the rendering performance in sparse aerial scenes.}
\label{tab:branch}
\end{table}

\noindent \textbf{NeRF Branch vs MPI Branch.}
In Table. \ref{tab:branch}, we examine the performance impact of the NeRF and MPI branches within our proposed MPNeRF. Initially, the NeRF branch without the multiplane loss $\mathcal{L}_{mul}$ (equals to a plain NeRF model) demonstrates a PSNR of 15.13, an SSIM of 0.20, and an LPIPS of 0.58. These values indicate a baseline level of performance where the NeRF branch struggles with sparse aerial views, as evidenced by the low PSNR and SSIM scores, along with a high LPIPS value which suggests a significant perceptual difference from the ground truth. In contrast, the MPI branch alone shows better across all metrics, with a PSNR of 20.32, an SSIM of 0.57, and a reduced LPIPS of 0.34. The MPI branch's improved performance is likely due to its discrete depth-based representation that aligns better with the structured nature of aerial scenes, thus capturing the scene geometry more effectively. And the inductive bias of CNN and Transformer makes MPI generalize better. The most significant performance gains are observed when the NeRF branch is combined with the multiplane loss $\mathcal{L}_{mul}$, resulting in a PSNR of 21.72, an SSIM of 0.80, and an LPIPS of 0.19. The addition of $\mathcal{L}_{mul}$ to the NeRF branch enhances its ability to recover details from sparse views, as reflected by the substantial improvements in PSNR, SSIM, and LPIPS. The proposed Multiplane Prior serves as a bridge to convey information that is hard to learn by the traditional NeRF pipeline. These results underscore the efficacy of incorporating multiplane priors into the NeRF framework for few-shot aerial scene rendering. 

\begin{figure*}
    \centering
    \includegraphics[width=1\linewidth]{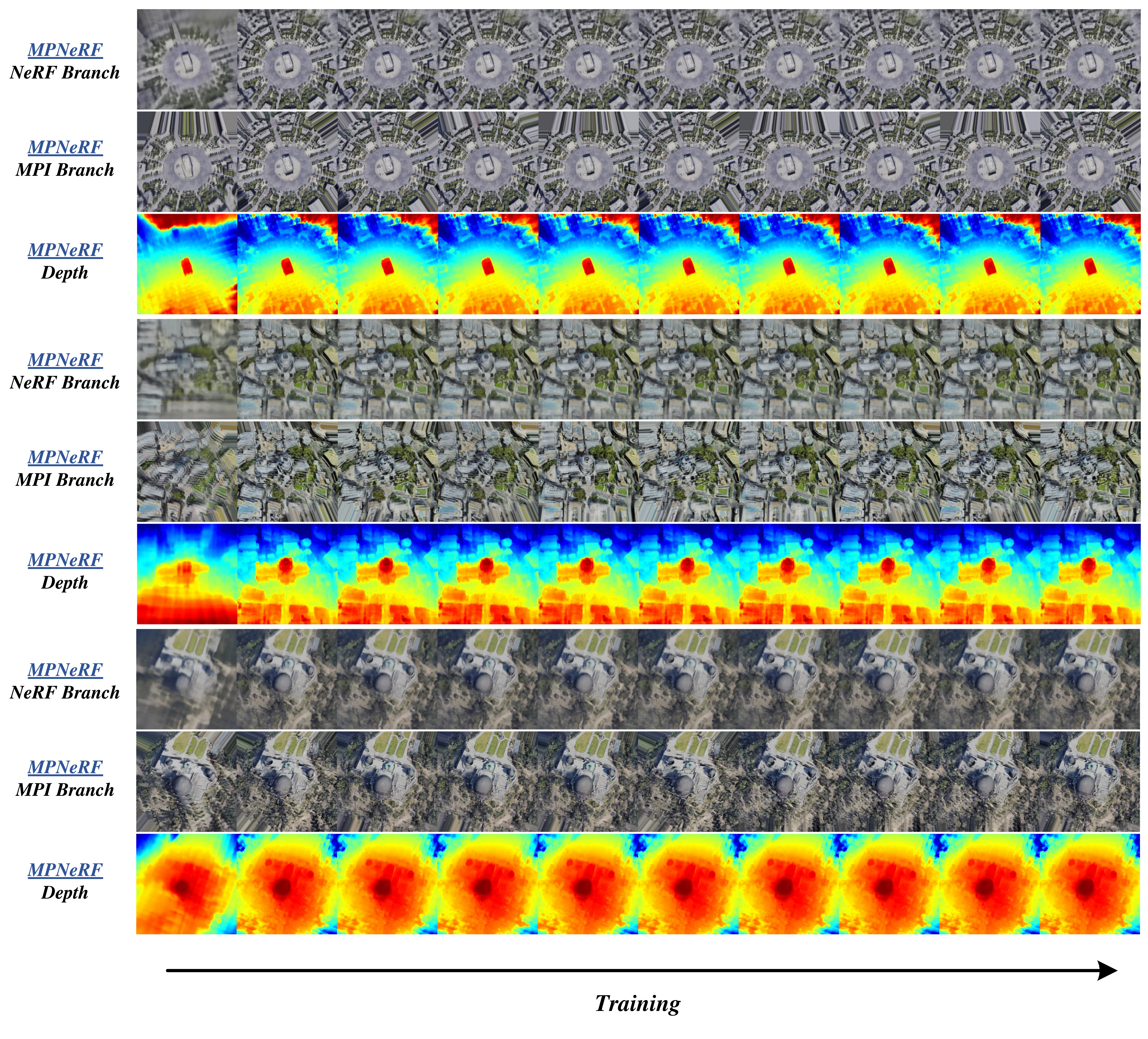}
    \caption{\textbf{Training Progression of MPNeRF.} The sequence shows comparative results from the NeRF and MPI branches at various training stages under 3 view settings. Left to right: early, mid, and late phases of training. The NeRF branch initially shows noisier reconstructions with indistinct depth estimations, while the MPI branch exhibits crop edge and overlapping ghosting effects. Over time, the NeRF branch, guided by the MPI-derived multiplane prior, progressively captures finer details and more accurate depth information, as reflected in the sharpening of depth map visualizations.}
    \label{fig:training}
\end{figure*}
\begin{figure}
    \centering
    \includegraphics[width=1\linewidth]{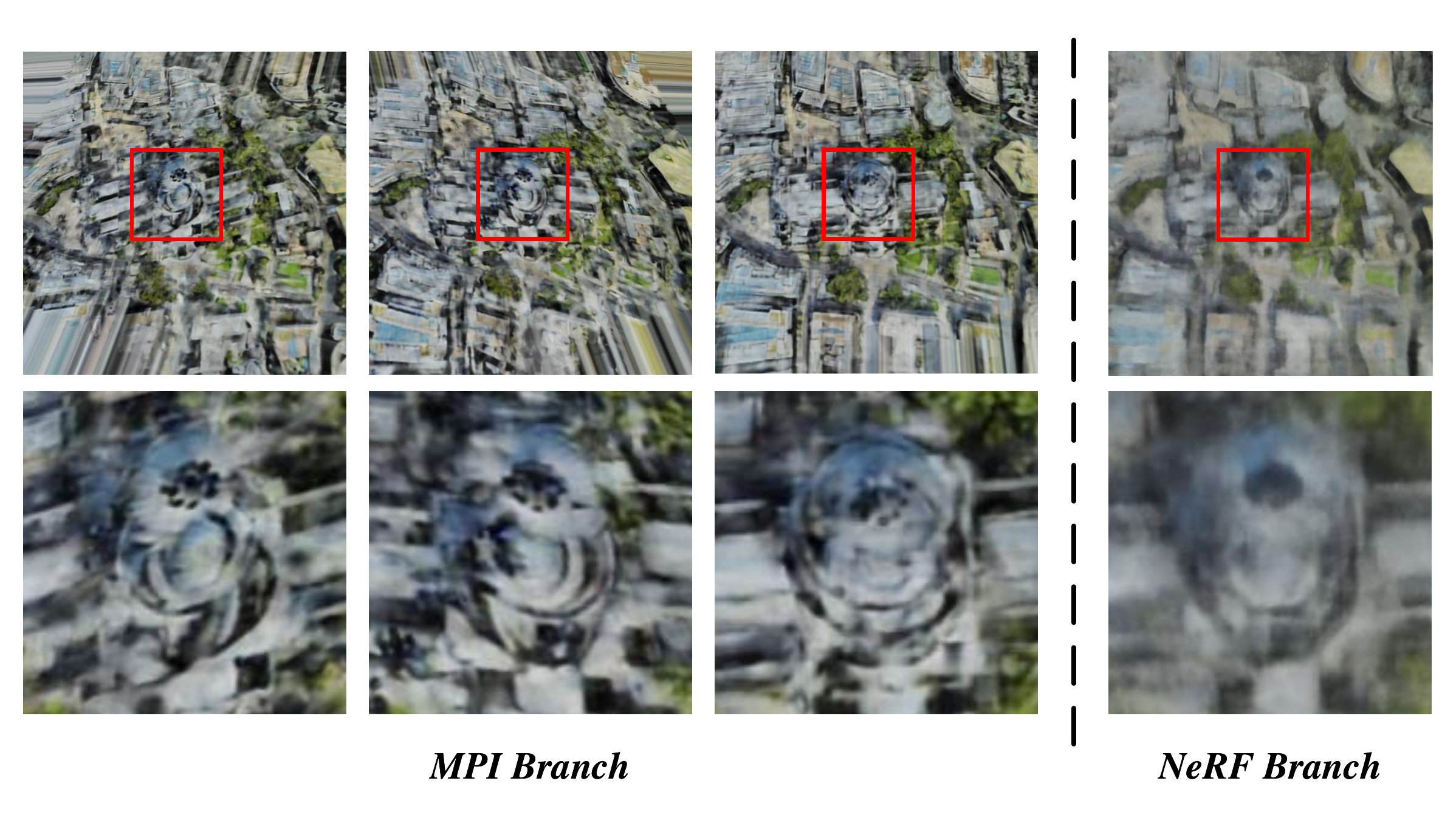}
    \caption{\textbf{Detail Comparison between MPI and NeRF Branch Outputs.} The images on the left column represent the MPI branch's output, displaying sharper details with overlapping ghosting effects in the highlighted regions. In contrast, the right column shows the NeRF branch's output, where the same regions appear more blurred. }
    \label{fig:mpi}
\end{figure}

\noindent \textbf{Other Few-shot NeRF Methods Combined with Multiplane Prior.} 
It stands to reason that it's worth evaluating other Few-shot NeRF methods combined with multiplane prior. We incorporated FreeNeRF's \cite{yang2023freenerf} frequency regularization and evaluated it under a 3-view setting. This integration results in a marginal increase in the PSNR by 0.2db. We believe the MPI's noisy predictions help reduce early training overfitting in high-frequency details. This mechanism seems to parallel the underlying concept of FreeNeRF, potentially explaining the minimal improvement. 
% We 

\section{Discussion and Future Works}
\noindent \textbf{Why MPNeRF Works?} Despite the advantage of the MPI representation in aerial scenes, a simple question is: Why MPNeRF is kept away from the cropped edge and overlapping ghosting effect of the MPI? Avoiding the cropped edge is simple, we sample rays from unseen views following the mask generated during homography warping. To better illustrate why the overlapping ghosting effect can not be learned by NeRF, we visualize the same target view rendered by the MPI branch in Figure. \ref{fig:mpi}. With the source viewpoint varying, the overlapping ghosting effect in the rendered target view differs. 
Since MPI derived from different viewpoints does not share a common world space, these overlapping ghosting effects are not multi-view consistent across all views. Thus these effects violate the multi-view consistency assumption of NeRF \cite{mildenhall2020nerf}. With these noises provided as pseudo-supervision, the MLP optimized with gradient descent tends to give blurry rendering. The blurring signifies NeRF's attempt to average out the incongruities across views. These pseudo-labels, while derived from an informed place, act as imperfect guides, introducing a trade-off that MPNeRF must navigate. On one hand, they provide a rich, albeit noisy, signal that captures the complexity of aerial scenes. On the other, they present a risk of polluting the training process with artifacts.

Although MPNeRF shows that a simple $\text{MSE}$ loss can perform well, this delicate balance highlights the importance of a carefully crafted training regimen, one that can differentiate between useful signals and misleading noise. Our future work will delve into refining this balance, potentially through the development of more sophisticated noise-filtering mechanisms or through the implementation of more robust training strategies that can better leverage the nuanced information within these pseudo-labels. In doing so, we may further enhance MPNeRF's rendering quality, pushing the boundaries of few-shot aerial scene rendering.

\noindent \textbf{Semantic Integration for Improved Scene Understanding.} Integrating semantic segmentation into the MPNeRF framework offers an exciting direction for enhancing scene understanding. By associating semantic labels with the MPI branch, MPNeRF may provide more contextually aware reconstructions and pave the way for applications in urban planning and navigation under limited data.

\noindent \textbf{Scene Editing.}
An exciting avenue for future research is the possibility of editing NeRF-rendered scenes by directly manipulating the MPIs generated by the MPI branch. This could enable users to alter scene characteristics such as color, texture, or even geometric structure, through an intuitive interface. A potential direction is utilizing differentiable rendering techniques to backpropagate the desired edits from the scene rendering back to the MPI and NeRF representations.

\noindent \textbf{Scalability.}
Currently, scalability remains a potential limitation when our MPNeRF model is applied to larger scenes. The primary bottleneck arises from the inherent capacity constraints of NeRF models. They are typically optimized for smaller, more controlled environments and can struggle to maintain fidelity at the increased scale and complexity of larger scenes. As scenes expand in size, the NeRF's neural network requires a corresponding increase in capacity to model the additional detail, which can lead to a significant escalation in computational and memory requirements \cite{tancik2022block, xu2023grid}. Furthermore, the encoder-decoder architecture employed within our MPI branch is not ideally suited for high-resolution imagery \cite{araujo2019computing, chen2019collaborative}. It tends to consume substantial amounts of memory, especially when processing the finer details necessary for large-scale scene rendering. The memory footprint grows rapidly with the resolution of input images due to the quadratic increase in the number of pixels that need to be processed simultaneously.

\section{Additional Visualizations.}

\noindent \textbf{Training Progression of MPNeRF.}
Figure. \ref{fig:training} presents a detailed visual account of the training evolution within our MPNeRF, delineating the comparative outcomes from the NeRF and MPI branches across three distinct training phases. The left columns illustrate the initial stage where the NeRF branch outputs are notably noisier, and the depth maps lack precise definition, signifying the model's initial struggle to interpret the sparse aerial views. These preliminary results are characterized by a lack of clarity and detail, with the depth maps displaying broad, undifferentiated regions of low confidence.
As training progresses to the midpoint, displayed in the center columns, the MPI branch starts to assert its strengths. It delivers reconstructions with improved clarity and begins to better capture the geometric intricacies of the aerial scenes. This enhancement is evident in the depth maps, where we observe a transition from broad, undefined areas to more distinct regions of depth estimation, indicative of the MPI branch's capability to delineate structural features more effectively at this stage.
Reaching the later stages of training, shown in the right columns, the NeRF branch, now informed by the multiplane prior, shows significant advancement. It starts to match and, in certain aspects, surpasses the MPI branch's performance by delivering images with greater detail fidelity. This is most apparent in the depth maps, where the once diffused and expansive high-confidence regions have now evolved into sharply defined areas, highlighting the network's improved proficiency in depth perception.

The visualization of the depth maps is particularly telling; the sharpening of these maps directly correlates with the improved model's depth estimations. The NeRF branch, leveraging the multiplane prior, demonstrates an enhanced ability to resolve the complex spatial relationships inherent in aerial scenes, moving beyond the initial limitations evidenced in the early training outputs.

This sequential improvement underscores the efficacy of the MPNeRF training process, which effectively leverages the distinct advantages of both NeRF and MPI branches to progressively refine the model's understanding of the scene, culminating in high-quality renderings from sparse inputs. The journey from noisy, indistinct initial attempts to clear, detailed final outputs exemplifies the potent potential of MPNeRF for aerial scene rendering.

\begin{figure*}
    \centering
    \includegraphics[width=1\linewidth]{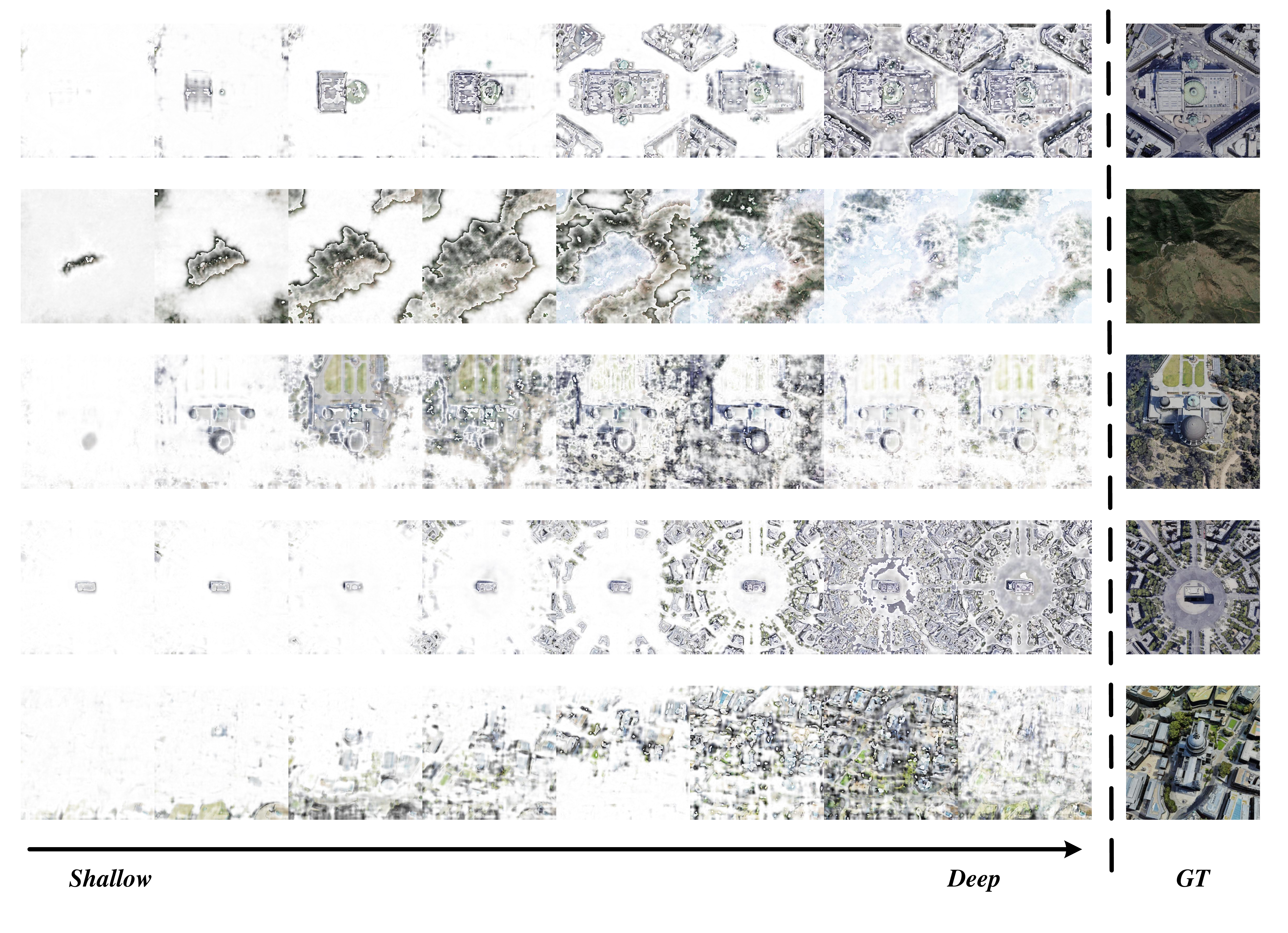}
    \caption{\textbf{Visualization of MPI Branch Depth Layers.} Sequential depth layers from the MPI branch reveal the aerial scene's structure, evolving from translucent to opaque as we move from shallow to deep layers, culminating in the ground truth (GT) image for reference.}
    \label{fig:mpi_layer}
\end{figure*}
\noindent \textbf{Different Layers of the MPI Branch.} To explore the geometry and appearance captured by the MPI branch, we visualize the color with transparency computed by the density of different MPI layers. Figure. \ref{fig:mpi_layer} showcases a series of images that represent different layers of the MPI branch, each corresponding to a specific depth level within the aerial scene, as labeled from 'Shallow' to 'Deep'. The images progress from the topmost layers, which capture high-elevation features like roofs, to the bottom layers, which reveal ground-level details. However, it is evident that the fidelity of the reconstruction varies across depth layers. The initial layers, while capturing the broad layout, lack the finer details and the sharpness present in the ground truth (GT). The middle layers begin to show more structure and texture, indicating an intermediate range where the MPI branch most effectively captures the scene's appearance. The deeper layers, while richer in detail, start to exhibit artifacts, such as blurring and possible misalignments, before converging towards the ground truth. This suggests that while the MPI branch of MPNeRF shows promise in reconstructing aerial scenes from limited data, it is still highly inaccurate and contains artifacts.

\section{More Implementation Details.}

\definecolor{tableheader}{rgb}{0.4,0.6,0.8}

\begin{table*}[]
\centering
\resizebox{\textwidth}{!}
{
\small
\begin{tabular}{@{} c | c | c | c | c | c @{}}
\toprule
\textbf{Layer} & \textbf{Kernel Size} & \textbf{In-Channels} & \textbf{Out-Channels} & \textbf{Input} & \textbf{Activation} \\
\midrule
convdown1 & 1 & 768        & 512          & encoder\_layer4                         & ELU \cite{clevert2015fast}                                     \\
convdown2 & 3 & 512         & 256          & convdown1                    & ELU                                      \\
convup1\_extra   & 3 & 256         & 256          & convdown2                    & ELU                                      \\
convup2\_extra   & 1 & 256         & 768         & convup1\_extra                      & ELU                                      \\
\midrule
convup5     & 3 & 768 + 21   & 256          & cat(convup2\_extra, depth\_embedding) & ELU                                      \\
conv5      & 3 & 256 + 768 + 21         & 256          & cat(convup5, encoder\_layer3, depth\_embedding)               & ELU                                      \\
convup4     & 3 & 256         & 128          & conv5                         & ELU                                      \\
conv4      & 3 & 128 + 384 + 21        & 128          & cat(convup4, encoder\_layer2, depth\_embedding)               & ELU                                      \\
output4     & 3 & 128         & 4            & conv4                         & Sigmoid (for RGB) and abs (for $\sigma$) \\
convup3     & 3 & 128         & 64           & conv4                         & ELU                                      \\
conv3      & 3 & 64 + 192 + 21          & 64           & cat(convup3, encoder\_layer1, depth\_embedding)               & ELU                                      \\
output3     & 3 & 64          & 4            & conv3                         & Sigmoid (for RGB) and abs (for $\sigma$) \\
convup2     & 3 & 64          & 32           & conv3                         & ELU                                      \\
conv2      & 3 & 32 + 96 + 21         & 32           & cat(convup2, encoder\_conv1, depth\_embedding)               & ELU                                      \\
output2     & 3 & 32          & 4            & conv2                         & Sigmoid (for RGB) and abs (for $\sigma$) \\
convup1     & 3 & 32          & 16           & conv2                         & ELU                                      \\
conv1      & 3 & 16          & 16           & convup1                        & ELU                                      \\
output1     & 3 & 16          & 4            & conv1                         & Sigmoid (for RGB) and abs (for $\sigma$) \\
\bottomrule
\end{tabular}
}
\caption{\textbf{Decoder Architecture for the MPI Branch.} Each convup layer within our architecture is composed of a convolution layer, followed by batch normalization and the specified activation layer, as delineated in the table. This sequence is then succeeded by a $2\times$ nearest neighbor upsampling process. Conversely, the convdown blocks are structured beginning with a max pooling layer with a stride of 2, followed by a convolution layer, and culminating with an activation layer. This architecture choice follows previous research in MPI representations and depth estimation \cite{li2021mine, wu2022remote, godard2019digging}.} \label{tab:arch}
\end{table*}
\noindent \textbf{Datasets and Metrics.} 
Our evaluation is conducted on a dataset that presents a rich tapestry of aerial landscapes, the LEVIR-NVS \cite{wu2022remote}, comprising 16 diverse scenes that span mountains, urban centers, villages, and standalone architectural structures. Each scene in the dataset is represented by a collection of 21 multi-view images, each with a resolution of $512 \times 512$ pixels. This selection ensures a broad representation of scenarios that MPNeRF might encounter in real-world applications. The LEVIR-NVS dataset encapsulates a variety of pose transformations that mimic the dynamic nature of UAV flight patterns, including wrapping and swinging motions. These pose variations introduce realistic challenges in aerial photography, such as changes in viewpoint and scale, making the dataset a rigorous testing ground for our model. The inclusion of these complex transformations in the simulation process is crucial for assessing the robustness of MPNeRF's performance in conditions that closely approximate actual aerial image capture. 

In our experimental setup, we strategically select specific views for training to assess the capability of our model in both interpolation and extrapolation scenarios. For the three-view setting, we utilize view IDs: 0, 7, and 15. This selection is designed to provide a spread of perspectives that challenges the model to extrapolate the scene effectively. In the five-view setting, we expand our selection to include view IDs: 0, 7, 10, 15, and 20. This broader range tests the model's interpolation skills and its ability to extrapolate scenes from more diverse viewpoints. 

In our experiments, we employ three standard metrics. Peak Signal-to-Noise Ratio (PSNR) is used to measure the image reconstruction quality, calculated as the negative logarithm of the mean squared error between the predicted and ground truth images. Structural Similarity Index Measure (SSIM), obtained via the skimage\footnote{https://scikit-image.org/} library, assesses image quality based on luminance, contrast, and structural information. Learned Perceptual Image Patch Similarity (LPIPS), computed using a VGG-based model from the lpips\footnote{https://github.com/richzhang/PerceptualSimilarity} package, evaluates perceptual similarity, reflecting more human-centric assessments of image quality.

\noindent \textbf{Implementation of Baseline Methods.}
We implement the baseline methods following their open-source code base. We adopt 64 coarse sampling and 32 fine sampling for the NeRF backbone of these methods. In particular, the RegNeRF \cite{niemeyer2022regnerf} and FreeNeRF \cite{yang2023freenerf} are implemented based on Mip-NeRF \cite{barron2021mip}, and others \cite{mildenhall2020nerf, kim2022infonerf, jain2021putting, yu2021pixelnerf} are based on a vanilla NeRF. All methods are trained for 30 epochs for each scene and the hyperparameters are strictly consistent across all experiments.  

\noindent \textbf{Implementation of MPNeRF.}
We implement MPNeRF based on the nerf-pl codebase \footnote{https://github.com/kwea123/nerfpl}, which provides a PyTorch Lightning framework for efficiently operationalizing NeRF architectures. The settings of hyperparameters are strictly consistent with baseline methods. Our NeRF branch adheres closely to the original NeRF paper specifications, ensuring a faithful reproduction of the baseline model. We adopt 64 coarse sampling and 32 fine sampling for the NeRF branch. Inspired by previous works \cite{tucker2020single, wu2022remote, li2021mine}, our MPI branch is constructed following an encoder-decoder architecture MPI generator. The encoder is a strand SwinV2 Transformer \cite{liu2022swin} pretrained via SimMIM \cite{xie2022simmim}. The encoder is kept frozen during training. A detailed description of our decoder architecture is presented in Table. \ref{tab:arch}. The MPI generator embeds depth hypotheses into the input features, which are then processed through convolutional layers to output MPIs with RGB and density values, leveraging skip connections and multi-scale representations for detail enhancement.

For optimization, we utilize the Adam optimizer \cite{kingma2014adam} with a learning rate of $5 \times 10^{-4}$, and a cosine learning rate decay scheduler \cite{loshchilov2016sgdr}. Our model is trained on a single NVIDIA RTX 3090 GPU for 30 epochs, taking about 2.5 hours to converge. The batch size is set to 1024 rays per iteration for both seen and unseen views, allowing sufficient diversity of data points for gradient estimation while maintaining manageable memory requirements. 
 \fi

\end{document}